\documentclass{article} 
\usepackage{iclr2027_conference,times}

\usepackage{amsmath,amsfonts,bm}

\def\eqref#1{equation~\ref{#1}}

\def\1{\bm{1}}

\DeclareMathAlphabet{\mathsfit}{\encodingdefault}{\sfdefault}{m}{sl}
\SetMathAlphabet{\mathsfit}{bold}{\encodingdefault}{\sfdefault}{bx}{n}

\usepackage{hyperref}
\usepackage{url}
\usepackage{booktabs}   
\usepackage{xcolor}
\usepackage{colortbl}   
\definecolor{groupgray}{gray}{0.9}
\definecolor{ours}{RGB}{232,240,254}   
\newcommand{\gain}[1]{\textcolor{green!45!black}{$+$#1}}
\usepackage[most]{tcolorbox}   
\usepackage{fvextra}           
\tcbset{promptbox/.style={colback=white, colframe=gray!70!black, coltitle=white, fonttitle=\bfseries, center title, rounded corners, boxrule=0.6mm, width=\linewidth, breakable, enhanced, left=6pt, right=6pt, top=4pt, bottom=4pt}}
\usepackage{pifont}            
\newcommand{\cmark}{\textcolor{green!45!black}{\ding{51}}}
\newcommand{\xmark}{\textcolor{red!70!black}{\ding{55}}}
\newcommand{\harn}[1]{\textcolor{red!70!black}{#1}}
\newcommand{\good}[1]{\textcolor{green!40!black}{#1}}
\tcbset{casebox/.style={enhanced, sidebyside, sidebyside align=top seam, sidebyside gap=10pt, lefthand width=0.47\linewidth, colback=white, colframe=gray!70!black, colbacktitle=gray!15, coltitle=black, fonttitle=\scriptsize, boxrule=0.5pt, arc=2pt, left=4pt, right=4pt, top=2pt, bottom=2pt, toptitle=1pt, bottomtitle=1pt, segmentation style={solid, gray!40}, before upper=\raggedright, before lower=\raggedright}}
\usepackage{multirow}   
\usepackage{makecell}   
\usepackage{enumitem}   
\usepackage{amsmath}    
\usepackage{graphicx}

\definecolor{hlgray}{HTML}{D0D8D0}
\definecolor{hlred}{HTML}{FF9994}
\definecolor{hlgreen}{HTML}{B5DC94}

\title{EvoIn: Bridging Evolution and Internalization for Agent Fine-Tuning}

\author{Shihan Dou* \quad Shaofan Liu* \quad Zhonghang Lu\thanks{The first three authors contributed equally.} \quad Jiahang Lin \quad Shichun Liu\\[1em]
\textbf{Binghai Wang} \quad \textbf{Jiajie Jin} \quad \textbf{Guanting Dong} \quad \textbf{Tao Gui} \quad \textbf{Qi Zhang} \quad \textbf{Xuanjing Huang}\\[1em]
Fudan University \quad Renmin University of China\\[0.3em]
\texttt{shihandou@foxmail.com, sfliu24@m.fudan.edu.cn, tgui@fudan.edu.cn}
}

\iclrfinalcopy 
\begin{document}

\maketitle

\begin{abstract}

Recent work has explored improving agents by jointly evolving their harnesses and models, but often takes a ``potpourri'' approach that bundles together new tools, new decision-making procedures, and model adaptation to the evolved harness under a single notion of agent improvement. 
In this paper, we instead investigate how agents can improve their decision-making procedures. 
In particular, we propose \textbf{EvoIn}, an agent fine-tuning framework that bridges \textbf{evo}lution and \textbf{in}ternalization. 
EvoIn first analyzes agent execution traces to evolve and validate new decision-making procedures by temporarily instantiating them in the harness. 
The validated procedures guide the agent to generate improved reasoning traces.
These traces are then rewritten into self-contained reasoning traces, removing explicit references to harness instructions while expressing the induced decision logic as the model's own reasoning. 
Finally, EvoIn fine-tunes the model on the rewritten traces, internalizing these procedures so that the improved decision-making persists without the evolved harness at inference time.
We evaluate EvoIn on diverse benchmarks and find that it consistently enables agents to learn stronger decision-making procedures, raising the pass rate by 10.9 points in-domain and by 9.2 points out-of-domain.
Results further show that the internalized decision procedures generalize to unseen tasks.
Case studies show that agents can learn to decide how to solve a task before solving it, for example by checking a document's length to choose between reading it in full and searching it.
EvoIn is also broadly applicable, showing consistent improvements on another model family.

\footnote{The source code for this manuscript is available at \href{https://github.com/Fanshaoliu/EvoIn}{https://github.com/Fanshaoliu/EvoIn}.}

\end{abstract}

\section{Introduction}

Decision-making procedures are central to the reasoning ability of language agents \citep{balke2014agents,rao1998decision,sumers2023cognitive}.
They determine how an agent reasons and acts, shaping behaviors such as planning, tool use, and verification \citep{yao2022react,yang2024swe,dou2026agents}.
Consider an agent asked to fix a software bug. 
Even with access to the same tools, one agent may patch the apparent failure immediately, whereas another may trace the relevant code, reproduce the failure, gather evidence, and check for regressions.
A stronger agent may further adjust its planning and verification to the task, performing lightweight checks for a simple local change but broader verification for a risky cross-module modification.  
We view an agent as a model coupled with a harness, where the harness denotes the model-external system that mediates the model's execution and interaction with the environment \citep{lin2026agentic,ning2026code}.
Decision-making procedures can be specified explicitly in the harness, through prompts, workflows, hooks, or other control logic \citep{lin2026agentic,yao2023tree,shinn2023reflexion}, or carried implicitly by the model and instantiated through its own reasoning \citep{chen2023fireact,qiao2024autoact}.
This raises natural questions: \textit{how can we improve agents' decision-making procedures, and to what extent can agents themselves drive or even automate this improvement process?}

\begin{figure}[t]
  \centering
  \includegraphics[width=0.9\linewidth]{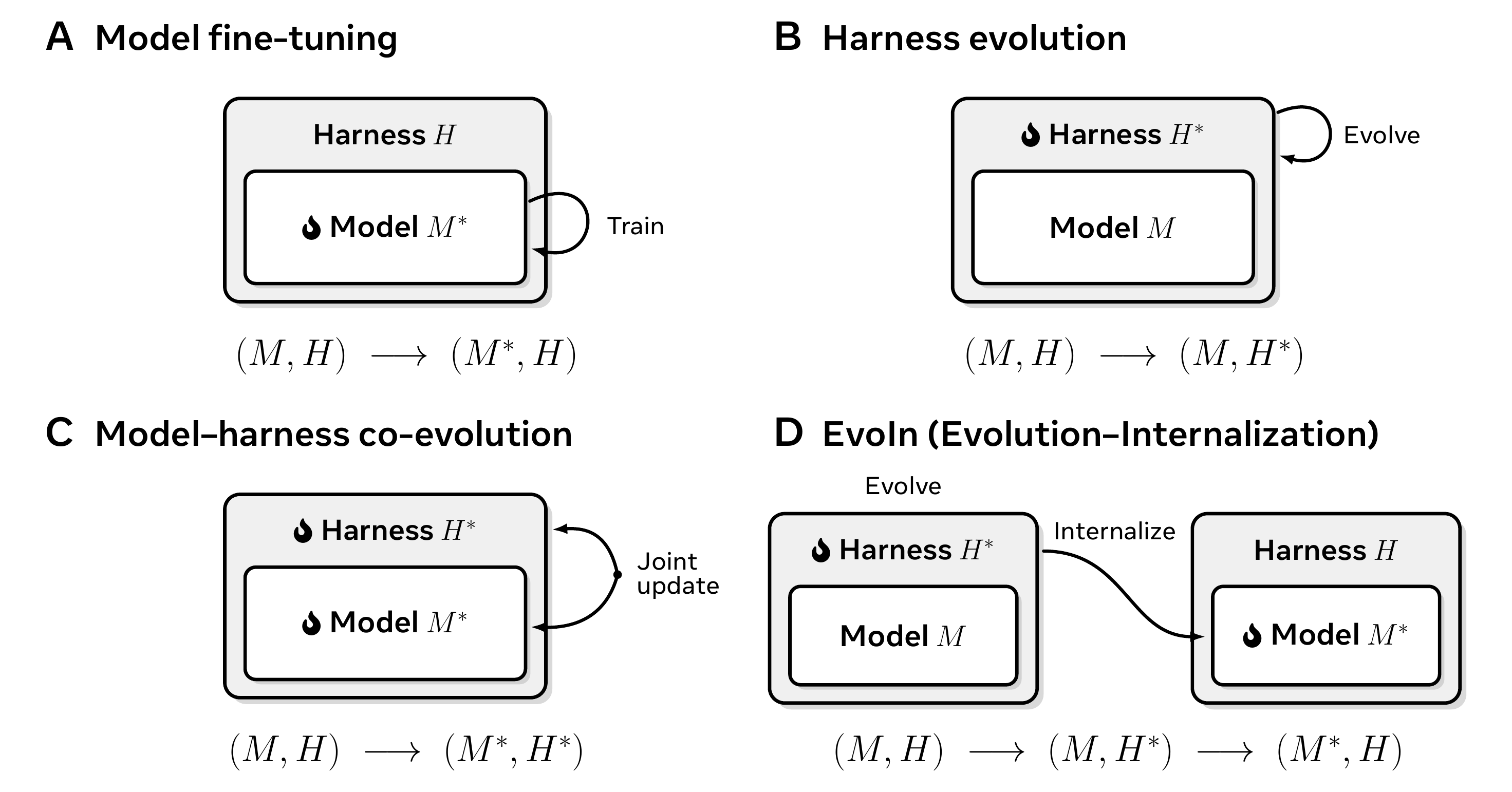}
\caption{Four paradigms for improving language agents.
\textbf{(A)} Model fine-tuning updates the model while keeping the harness fixed.
\textbf{(B)} Harness evolution updates the harness while keeping the model fixed.
\textbf{(C)} Model--harness co-evolution updates both jointly.
\textbf{(D)} EvoIn first evolves the harness to discover improved decision procedures, then internalizes the reusable procedures into the model and returns to the original harness.
Flame icons and superscript $*$ indicate updated components.}
  \label{fig:agent-paradigms}
\end{figure}

We compare existing paradigms for improving agents' decision-making procedures, as illustrated in Figure~\ref{fig:agent-paradigms}.
Model fine-tuning approaches keep the harness fixed and improve the language model, typically through better tasks, environments, trajectories, or rewards \citep{chen2024agent,zeng2024agenttuning,fu2025agentrefine,jimenez2024swe}.
In this paradigm, agents may help construct training data or collect experience, but there is no explicit space to propose, test, and validate candidate procedure changes before training.
This also limits the exploration of decision procedures\footnote{We use \emph{decision-making procedure}, \emph{decision procedure}, and \emph{procedure} interchangeably in this paper.} that are easier to express and evaluate externally, especially those that may not be readily acquired through direct model updates.
Harness evolution takes the opposite route, keeping the model fixed while explicitly modifying the harness to change the decision-making procedures that govern how the agent solves tasks \citep{lin2026agentic,zhang2026self,lee2026meta}.
However, it alone leaves these improvements external to the underlying language model, without a chance to internalize reusable decision procedures as stronger model reasoning capabilities.

More recent work co-evolves models and harnesses, combining the two directions above \citep{chen2026co,lee2026recursive,chen2026harnessforge,chen2026harnessx}. 
While effective for optimizing the final agent system, this jointly rewards new tools, decision procedures, and model adaptation under the same end-to-end objective, making it difficult to tell whether the model has learned stronger decision-making procedures or simply become better matched to the evolved harness \citep{yu2026coevolvingharnessesmodels}.
More importantly, training the model with the evolved harness confounds stronger reasoning with harness-specific adaptation, making it unclear whether the decision logic learned by the model during training transfers and generalizes across different harnesses.

In this work, we propose \textbf{EvoIn}, an agent fine-tuning framework that bridges the explicit exploration enabled by harness \textbf{evo}lution and the lasting capability gains enabled by model \textbf{in}ternalization, while allowing agents to drive much of the process.
Starting from training data, the agent first analyzes the reasoning traces and answers produced under the current harness, proposes improvements to its decision-making procedures, and evaluates the resulting changes.
This process can be repeated for multiple rounds to progressively refine the procedures based on evaluation feedback.
The evolved harness then guides the generation of improved reasoning traces. Because these traces may explicitly refer to procedural instructions introduced by the evolved harness, they are rewritten into self-contained reasoning traces that preserve the induced decision logic while expressing it as the model's own reasoning. 
Finally, the rewritten traces are used with the original harness to fine-tune the model. 
In this way, the harness provides a temporary space for agents to discover and validate better decision procedures, which are ultimately learned by the model.

Concretely, EvoIn raises accuracy by 10.9 points on the in-domain categories and by 9.2 points on the out-of-domain benchmarks, with gains of up to 19.7 points on a single benchmark.
More specifically, rewriting is necessary for internalization: a model fine-tuned on the unrewritten trajectories cites instructions and tools that do not exist in the original harness (Figure~\ref{fig:cases}(c)), runs out of turns on 19.4\% of out-of-domain examples compared with 6.1\% for EvoIn, and keeps almost none of the out-of-domain gain.
The learned procedures also transfer to different task types, where the model can reason about unseen tasks and generate task-appropriate decision procedures rather than simply replay those encountered during evolution.
Experimental results on Gemma also show clear gains, suggesting that EvoIn generalizes across model families.
Moreover, case studies show that agents can discover and learn procedures that dynamically adapt to the input, such as checking the document length before choosing how to read the document (Appendix~\ref{app:case_length}).
In summary, with its simple fine-tuning pipeline and promising results, we hope EvoIn offers a practical path toward more capable agents with less human intervention and inspires future work toward agent self-improvement.

\section{Related Work}
\label{sec:related}

In this section, we position \textbf{EvoIn} relative to existing approaches for improving language agents.
We organize prior work according to which part of the agent is updated.
Formally, we view a language agent as a pair $A=(M_\theta,H_0)$, consisting of a model $M_\theta$ and its original harness $H_0$ (a.k.a. seed harness).
Decision procedures may reside implicitly in the model or be specified explicitly in the harness.
Under this view, model fine-tuning updates only the model, $(M_\theta,H_0)\rightarrow(M_{\theta^\star},H_0)$; harness evolution updates only the harness, $(M_\theta,H_0)\rightarrow(M_\theta,H^\star)$; and model--harness co-evolution updates both, $(M_\theta,H_0)\rightarrow(M_{\theta^\star},H^\star)$.
EvoIn follows a fourth path,
\begin{equation}
(M_\theta,H_0) \xrightarrow{\text{evolution}} (M_\theta,H^\star)
\xrightarrow{\text{internalization}} (M_{\theta^\star},H_0),
\label{eq:evoin}
\end{equation}
where the evolved harness serves as a temporary space for proposing and validating improved decision procedures before they are internalized.
Figure~\ref{fig:agent-paradigms} summarizes these four paradigms.

\textbf{Model fine-tuning $(M_\theta,H_0)\to(M_{\theta^\star},H_0)$.}
This path fixes the harness and pushes all improvement into weights.
Instruction tuning and preference alignment shape general behavior \citep{ouyang2022training,rafailov2024directpreferenceoptimizationlanguage}, agentic data synthesis supplies multi-turn supervision from tool-call annotations, trajectories, and executable environments \citep{schick2023toolformer,zeng2024agenttuning,pan2025trainingsoftwareengineeringagents}, and reinforcement learning with verifiable rewards optimizes the policy against task outcomes \citep{Guo_2025,jin2025searchr1trainingllmsreason,dong2025agenticreinforcedpolicyoptimization}.
Throughout, the improvement loop is designed by humans, since tasks, environments, rewards, and algorithms are given, and the agent at most produces data and experience, never a proposal about how it should decide.
Decision-making procedures therefore change only as a byproduct of gradient updates, rather than existing as objects that can be stated, held fixed, and compared. Yet it is the capability gained along this path that makes it possible to hand the job of improving the agent to the agent itself.

\textbf{Harness evolution $(M_\theta,H_0)\to(M_\theta,H^\star)$.}
Once a model can reflect on its own trajectories, diagnose failures, and revise its behavior from feedback \citep{shinn2023reflexion,madaan2023self}, the harness becomes an object the agent operates on directly, and the scope it is allowed to rewrite has widened steadily, from prompts and context \citep{khattab2023dspycompilingdeclarativelanguage,agrawal2026gepareflectivepromptevolution}, to reusable skills and experiential memory \citep{wang2023voyager,zhao2024expel}, finally, to the module composition and code of the harness itself \citep{hu2025automated,novikov2025alphaevolvecodingagent,zhang2026darwin,lin2026agentic}.
This supplies exactly the space that model fine-tuning lacks, since a procedure here is explicitly written and executably verifiable, so it can be proposed, tested, kept, or discarded.
Its product, however, is always external.
What improves is the harness while $M_\theta$ is untouched, so a discovered procedure survives only as an instruction, a skill, or a piece of code that must be carried along to take effect, and never settles into the model's own reasoning.
This externality is costly in practice, as skill libraries degrade once they accumulate without lifecycle management \citep{zhang2026librarydriftdiagnosing}, harness evolution does not consistently beat test-time scaling under matched budgets and generalizes weakly to held-out tasks \citep{wang2026rethinkingevaluationharness}, and benefiting from a harness update is a capability distinct from producing one \citep{lin2026harnessupdatingnotharness}.

\textbf{Model and harness co-evolution $(M_\theta,H_0)\to(M_{\theta^\star},H^\star)$.}
A recent line optimizes both sides in one loop, alternating harness or skill search with weight updates under the searched configuration \citep{chen2026harnessforge,chen2026co,chen2026harnessx,lee2026recursive}.
These methods do update the model, but they also make what is written into it hard to identify, because the model is trained on trajectories generated under $H^\star$ and evaluated under $H^\star$ as well.
The coupling is concrete.
Training a weaker model on a stronger expert's trajectories under the weaker model's own evolved harness regresses performance on all seven tasks studied by 4 to 30 points, since the expert's planning style no longer matches the harness evolved around the weaker model \citep{yu2026coevolvingharnessesmodels}, and harness choice alone moves measured accuracy by as much as 28 points within a single model \citep{starace2026scaffoldeffectsgaia}.
Harness design and post-training thus interact \citep{kim2026interplayharnessdesign}, so a gain measured under $H^\star$ does not by itself show that the model's own decision-making has improved.
In contrast, when trajectories are rewritten for the original harness and the model is evaluated under it, trajectories from a stronger model under the evolved harnesses of a weaker one improve the weaker model (Section~\ref{sec:further_analysis}).

\textbf{Internalizing the harness $(M_\theta,H_0)\to(M_\theta,H^\star)\to(M_{\theta^\star},H_0)$.}
Closest to us is work that moves harness-side structure into the model, from context distillation \citep{askell2021general,snell2022learningdistillingcontext} to the internalization of prompted reasoning, tool use, and explicit chains of thought \citep{chen2023fireact,qiao2024autoact,yu2024distilling,deng2024explicit}, including recent work that treats the harness as a training-time teacher and removes it at inference \citep{dennis2026compiling,wu2026seed}.
However, in this line of work, the procedure to be internalized is typically specified in advance, such as a hand-designed strategy or skill.
EvoIn differs in three respects.
First, the procedure to be internalized is proposed and validated by an agent through iterative harness evolution rather than fixed beforehand.
Second, the agent rewrites the trajectories to remove evolved-harness dependencies while preserving the induced decision logic.
Third, the model is fine-tuned and evaluated under the original harness $H_0$, separating procedure learning from harness-specific adaptation.
These choices provide an opportunity for the entire improvement process to become agent-driven and allow agents to improve what is ultimately internalized into the model.

\begin{figure}[htpb]
  \centering
  \includegraphics[width=0.94\linewidth]{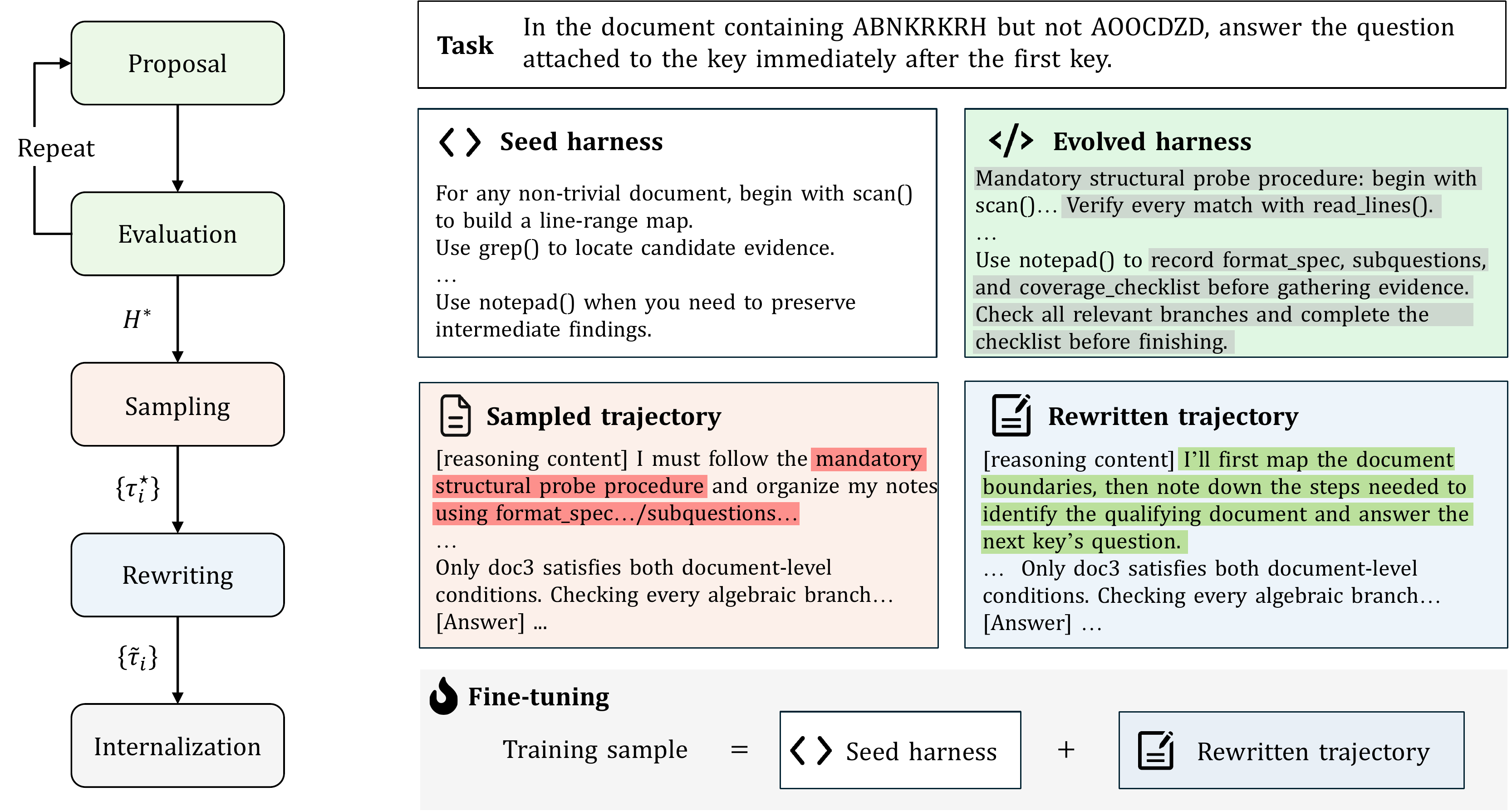}
  \caption{We illustrate EvoIn with a multi-document reasoning case.
EvoIn refines decision procedures through iterative proposal and evaluation, then samples and rewrites trajectories for fine-tuning.
\colorbox{hlgray}{Gray} highlights mark procedure updates in the evolved harness.
\colorbox{hlred}{Red} highlights show explicit reliance on instructions from the evolved procedures, while \colorbox{hlgreen}{green} highlights show the same decision logic reformulated as the agent's own reasoning.
  }
  \label{fig:method}
\end{figure}

\section{EvoIn: Bridging Evolution and Internalization}
\label{sec:method}

As illustrated in Figure~\ref{fig:method}, EvoIn consists of five stages: \textit{procedure proposal}, \textit{procedure evaluation}, \textit{procedure-guided sampling}, \textit{rewriting}, and \textit{procedure internalization}.
For a task $x$, let $\tau$ denote its execution trajectory, including the model's reasoning, tool calls and corresponding observations when applicable, and the final answer.
We use $z$ to denote the reasoning trace within $\tau$.
Given training data $\mathcal{D}_{\mathrm{train}}$ and held-out development data $\mathcal{D}_{\mathrm{dev}}$, EvoIn realizes the evolution--internalization process in Equation~\eqref{eq:evoin}.
A subset of $\mathcal{D}_{\mathrm{train}}$ is used during evolution for trajectory analysis, while $\mathcal{D}_{\mathrm{dev}}$ is used only to evaluate candidate harnesses.

\textbf{Procedure Proposal.}
Evolution starts from the original harness $H_0$, which is evaluated without modification in the initial round.
In each subsequent round, a proposal agent$^2$ examines the current harness together with feedback and execution trajectories from the previous round, and proposes candidate harness updates that instantiate improved decision procedures.
The update may revise the system prompt, workflow, or other control logic that specifies the decision procedure.
Each candidate must pass static and runtime validation before evaluation, with details provided in Appendix~\ref{app:pipeline_details}.
The proposal prompt is provided in Appendix~\ref{app:proposal_prompt}.

\textbf{Procedure Evaluation.}
This stage evaluates whether each proposed procedure is both effective and general.
Each candidate harness is paired with the same target model $M_\theta$ and evaluated on a held-out $\mathcal{D}_{\mathrm{dev}}$ using the task evaluator.
Meanwhile, an analysis agent examines successful and failed trajectories from the evolution subset of $\mathcal{D}_{\mathrm{train}}$ and summarizes failure patterns and actionable feedback for the next proposal round.
Proposal and evaluation are repeated for multiple rounds, and the highest-scoring harness on $\mathcal{D}_{\mathrm{dev}}$ across all rounds is retained as the evolved harness $H^\star$.

\textbf{Procedure-Guided Sampling.}
After the iterative proposal and evaluation process, the resulting harness $H^\star$ is used with the target model $M_\theta$ to collect trajectories on $\mathcal{D}_{\mathrm{train}}$.
A task evaluator scores the resulting trajectories, and only trajectories with the full task score and a valid final answer are retained for internalization.
We denote these procedure-guided trajectories by $\{\tau_i^\star\}$.
Their reasoning and actions reflect the improved decision procedures.

\textbf{Rewriting.}
Trajectories generated under $H^\star$ cannot always be used directly for fine-tuning under $H_0$.
Their reasoning may explicitly refer to instructions, tools, or runtime assumptions that exist only in the evolved harness, causing the fine-tuned model to reproduce such dependencies after $H^\star$ is removed.
For example, a reasoning trace may state that ``according to the instructions, I should first retrieve the database,'' even though no such instruction exists in the original harness $H_0$.
So we rewrite each trajectory into a self-contained trajectory that could have been naturally produced under the original harness while preserving the useful decision logic induced by $H^\star$.

We implement this stage with a pipeline called \textsc{Tailor}, which consists of three steps.
\textbf{First}, a rewriting phase edits each trajectory to remove dependencies on the evolved harness while preserving its useful decision logic.
This may involve rephrasing harness-specific reasoning, adapting unsupported tool calls to $H_0$, or removing unnecessary steps.
\textbf{Second}, the rewritten trajectory is replayed under $H_0$, followed by deterministic checks for issues such as invalid tool calls or residual harness-specific dependencies.
Trajectories that fail these checks undergo one targeted repair and are then replayed again.
\textbf{Finally}, another checking phase verifies that the final answer is supported by the visible evidence and that the trajectory is suitable for training.
Only trajectories that pass this check are retained.
We denote the resulting rewritten trajectories by $\{\tilde{\tau}_i\}$.
Appendix~\ref{app:pipeline_details} describes the replay and repair procedure, and Appendix~\ref{app:tailor_prompts} provides the corresponding prompts.

\textbf{Procedure Internalization.}
Finally, the original harness $H_0$ is restored and $M_\theta$ is fine-tuned on the rewritten trajectories using supervised fine-tuning.
The loss is applied to model-generated turns.
This produces $M_{\theta^\star}$ and returns the agent to $(M_{\theta^\star},H_0)$, so the improved decision procedures no longer depend on the evolved harness at inference time.

\section{Experiments}
\label{sec:experiments}

Our experiments test two claims: task-specific harness evolution produces better training demonstrations, and supervised fine-tuning (SFT) internalizes the resulting decision procedures.

\subsection{Setup}
\label{sec:experiment_setup}

\textbf{Dataset.}
The in-domain (ID) data comprise 23 task categories, which we organize into seven task groups (Appendix~\ref{app:id_categories}).
Harness evolution analyzes failures on an evolution split of each category and compares candidate harnesses on a disjoint evolution-test split ($\mathcal{D}_{\mathrm{dev}}$ in Section~\ref{sec:method}), which never enters the SFT corpus and later serves as the ID test set.
The training data come from $\mathcal{D}_{\mathrm{train}}$, the evolution split together with the remaining non-test examples: the best harness of each category rolls out the target model on these examples, a judge model keeps only fully correct trajectories, and Tailor (Section~\ref{sec:method}) rewrites, repairs, and checks them.
For out-of-domain (OOD) testing, we use six benchmarks: AA-LCR~\citep{artificialanalysis2025lcr}, BrowseComp-LongContext (adapted from BrowseComp, \citealp{wei2025browsecomp}), LongBench v2~\citep{bai2024longbench2}, MRCR~\citep{vodrahalli2024michelangelo}, Oolong~\citep{bertsch2025oolong}, and Table-Longer.
Appendix~\ref{app:ood_benchmarks} describes each benchmark's task type, and none appears among the recorded sources of evolution or SFT data.

\textbf{Seed harness.}
Every category starts from the same task-agnostic seed harness, which serves as the original harness $H_0$ and is the only harness used at evaluation.
The seed harness does not place the document in the prompt; instead, the agent reads it through six actions, one per turn: (1) \texttt{scan}, which returns a line-range map of the whole document; (2) \texttt{grep[pattern]}, which returns every matching line number with a short preview; (3) \texttt{read\_lines[start, end]}, which returns the raw lines in a range; (4) \texttt{bash[command]}, which runs a shell command, typically Python, over the document for arithmetic, counting, and sorting; (5) \texttt{notepad[action, content]}, which writes or reads a scratchpad that persists across turns; and (6) \texttt{complete\_task[result]}, which submits the final answer and ends the episode.
The system prompt asks the agent to support every claim with lines it has read or with a computation, the agent may take at most 30 turns, and long tool outputs are cut to their beginning and end (Appendix~\ref{app:pipeline_details}).
Evolved harnesses may rewrite these instructions and tool descriptions or raise the turn budget, but they are used only during evolution and trajectory collection, both of which happen before fine-tuning.

\textbf{Implementation details.}
Unless stated otherwise, all experiments use the following settings.
Qwen3.5-35B-A3B~\citep{qwen3.5} is the target model $M_\theta$; it performs every rollout, both during evolution and when collecting training trajectories, and is also the SFT student.
Claude Opus 4.7~\citep{anthropic2026opus47} acts as both the analysis agent and the proposal agent, and for each category it runs ten evolution rounds with three candidate harnesses per round (Appendix~\ref{app:proposal_prompt}).
Tailor uses GLM-5.2~\citep{glm5} as both the rewriting model and the checking model.
We fine-tune the target model with full-parameter SFT and report the final checkpoint, and all Qwen ablations use the same recipe (Appendix~\ref{app:sft}).
For the main run, we train on 12,035 demonstrations for three epochs, with a maximum sequence length of 24,576 tokens.
We report two metrics, computed from one sampled trajectory per example under a fixed protocol: for an example score $s\in[0,1]$, \emph{Pass} is the percentage of examples with $s=1$, and \emph{Score} is a rubric score, 100 times the mean example score, which gives partial credit where the rubric or official metric allows it.
OOD results pool the examples of all six benchmarks, so each benchmark is weighted by its number of examples.
Appendix~\ref{app:eval} lists the judges and decoding settings.

\subsection{Main Results}
\label{sec:main_results}

\textbf{EvoIn improves ID and OOD performance under the seed harness.}
Tables~\ref{tab:id_results} and~\ref{tab:ood_results} compare Qwen3.5-35B-A3B before and after EvoIn fine-tuning.
EvoIn raises Pass from 35.96\% to 46.83\% on the 2,300 ID examples ($+10.87$ points, a 30\% relative gain) and from 18.99\% to 28.20\% on the 2,064 OOD examples ($+9.21$ points, a 48\% relative gain).
Since both models run with the same seed harness and evaluation protocol, these gains come from the fine-tuned weights.

\begin{table}[t]
  \centering
  \caption{\textbf{ID results by task group} (\%).
  Base is Qwen3.5-35B-A3B, and EvoIn is the same model after EvoIn fine-tuning (epoch 3).
  Each group value is the mean over its categories, each with 100 evolution-test examples; Appendix~\ref{app:id_categories} lists the categories and their results.
  $\Delta$ is EvoIn minus Base, and Pass equals Score for groups whose categories all have binary scores.}
  \label{tab:id_results}
  \small
  \setlength{\tabcolsep}{5pt}
  \begin{tabular}{lrr>{\columncolor{ours}}rrr>{\columncolor{ours}}rr}
    \toprule
    & & \multicolumn{3}{c}{Pass} & \multicolumn{3}{c}{Score} \\
    \cmidrule(lr){3-5}\cmidrule(lr){6-8}
    Task group & $N$ & Base & EvoIn & $\Delta$ & Base & EvoIn & $\Delta$ \\
    \midrule
    Multi-document key retrieval & 300 & 57.00 & \textbf{63.33} & \gain{6.33} & 57.00 & \textbf{63.33} & \gain{6.33} \\
    Evidence-grounded QA & 400 & 52.25 & \textbf{61.50} & \gain{9.25} & 55.03 & \textbf{64.93} & \gain{9.90} \\
    Structured-data reasoning & 300 & 44.00 & \textbf{61.33} & \gain{17.33} & 44.00 & \textbf{61.33} & \gain{17.33} \\
    Log and dialogue tracking & 300 & 53.00 & \textbf{75.00} & \gain{22.00} & 55.14 & \textbf{76.78} & \gain{21.64} \\
    In-context learning & 200 & 26.00 & \textbf{39.00} & \gain{13.00} & 26.00 & \textbf{39.00} & \gain{13.00} \\
    Document-grounded generation & 300 & 22.00 & \textbf{36.00} & \gain{14.00} & 22.00 & \textbf{36.00} & \gain{14.00} \\
    Complex instruction following & 500 & 7.60 & \textbf{9.20} & \gain{1.60} & 43.75 & \textbf{49.41} & \gain{5.66} \\
    \midrule
    All 23 categories & 2,300 & 35.96 & \textbf{46.83} & \gain{10.87} & 44.58 & \textbf{56.40} & \gain{11.82} \\
    \bottomrule
  \end{tabular}
\end{table}

\textbf{EvoIn improves performance across all seven ID task groups.}
All seven task groups improve, and ID Score rises by 11.82 points overall (Table~\ref{tab:id_results}).
The largest gains are on log and dialogue tracking ($+21.64$ Score) and structured-data reasoning ($+17.33$), both require counting, tracking, or computing over many records.
Complex instruction following gains the least ($+5.66$), and its Pass stays below 10\% for both models.
At the category level, Score increases in 21 of the 23 categories and decreases in Needle QA ($-6.00$) and Long-Source Deliverables ($-1.77$), as detailed in Table~\ref{tab:app_id_results}.

\begin{table}[t]
  \centering
  \caption{\textbf{OOD results by benchmark} (\%).
  Models and $\Delta$ follow Table~\ref{tab:id_results}.}
  \label{tab:ood_results}
  \small
  \setlength{\tabcolsep}{5pt}
  \begin{tabular}{lrr>{\columncolor{ours}}rrr>{\columncolor{ours}}rr}
    \toprule
    & & \multicolumn{3}{c}{Pass} & \multicolumn{3}{c}{Score} \\
    \cmidrule(lr){3-5}\cmidrule(lr){6-8}
    Benchmark & $N$ & Base & EvoIn & $\Delta$ & Base & EvoIn & $\Delta$ \\
    \midrule
    AA-LCR & 100 & 27.00 & \textbf{37.00} & \gain{10.00} & 27.00 & \textbf{37.00} & \gain{10.00} \\
    BrowseComp-LongContext & 295 & 12.54 & \textbf{13.56} & \gain{1.02} & 12.54 & \textbf{13.56} & \gain{1.02} \\
    LongBench v2 & 503 & 33.20 & \textbf{44.73} & \gain{11.53} & 33.20 & \textbf{44.73} & \gain{11.53} \\
    MRCR & 800 & 2.00 & \textbf{8.88} & \gain{6.88} & 32.55 & \textbf{76.31} & \gain{43.76} \\
    Oolong & 300 & 43.33 & \textbf{60.33} & \gain{17.00} & 44.58 & \textbf{61.42} & \gain{16.84} \\
    Table-Longer & 66 & 22.73 & \textbf{42.42} & \gain{19.69} & 22.73 & \textbf{42.42} & \gain{19.69} \\
    \midrule
    Average & 2,064 & 18.99 & \textbf{28.20} & \gain{9.21} & 31.02 & \textbf{54.49} & \gain{23.47} \\
    \bottomrule
  \end{tabular}
\end{table}

\textbf{The gains of EvoIn also transfer to all six OOD benchmarks.}
EvoIn improves all six OOD benchmarks in both Pass and Score, and its OOD Pass gain is close to its ID gain (Table~\ref{tab:ood_results}).
Score gains range from $+1.02$ on BrowseComp-LongContext to $+43.76$ on MRCR, and the other four benchmarks gain between 10.00 and 19.69 points.
Overall Score rises by 23.47 points, much more than Pass, and almost all of this difference comes from MRCR, whose official metric gives partial credit and whose Pass rises by only 6.88 points.

\section{Further Analysis}
\label{sec:further_analysis}

This section examines where the gains of EvoIn come from and how far they extend.
We test whether harness evolution and rewriting are both needed, whether the target model can drive the pipeline by itself, whether a stronger rollout model provides better demonstrations, and whether EvoIn works for another model family.
Appendix~\ref{app:ablation_details} describes how each run is constructed and gives its per-category and per-benchmark results.

\begin{table}[t]
  \centering
  \caption{\textbf{Ablation study} (\%).
  All rows use the same Qwen student, SFT schedule, seed harness, and evaluation protocol as Tables~\ref{tab:id_results} and~\ref{tab:ood_results}. Corpus size counts SFT demonstrations.
  The top block separately ablates harness evolution and rewriting.
  The bottom block compares EvoIn using equally sized Qwen and GLM-5.3 rollout subsets from the same tasks. Bold marks each block's best result.
}
  \label{tab:component_ablation}
  \small
  \setlength{\tabcolsep}{5pt}
  \begin{tabular}{lrrrrr}
    \toprule
    & & \multicolumn{2}{c}{ID} & \multicolumn{2}{c}{OOD} \\
    \cmidrule(lr){3-4}\cmidrule(lr){5-6}
    Method & Corpus size & Pass & Score & Pass & Score \\
    \midrule
    Base & 0 & 35.96 & 44.58 & 18.99 & 31.02 \\
    Seed-harness SFT & 12,035 & 37.83 & 47.25 & 18.36 & 43.16 \\
    EvoIn w/o Tailor & 12,035 & 45.00 & 54.89 & 19.67 & 45.70 \\
    \rowcolor{ours} EvoIn & 12,035 & \textbf{46.83} & \textbf{56.40} & \textbf{28.20} & \textbf{54.49} \\
    \midrule
    Qwen rollout & 1,752 & 46.52 & 56.30 & 22.04 & 49.36 \\
    GLM-5.3 rollout & 1,752 & \textbf{52.43} & \textbf{61.24} & \textbf{25.68} & \textbf{49.65} \\
    \bottomrule
  \end{tabular}
\end{table}

\begin{figure}[t]
\centering
\scriptsize
\begin{tcolorbox}[casebox, title={\textbf{(a) Harness evolution: the evolved harness makes the base model record, sort, and recheck every row.}}]
\textbf{Seed harness} \xmark\par
\texttt{read\_lines(4, 13)} returns all nine WS4 rows.\par
\textit{``\dots I'll return only those who actually played at 2nd base.''}\par
\textit{``\dots holliho01 and whitewa01, each with 9 games.''} (9 is whitewa01's g\_3b)\par
Answer: \texttt{holliho01 9; whitewa01 1} \harn{(seven rows missing; whitewa01 has 0)}
\tcblower
\textbf{Evolved harness} \cmark\par
\texttt{read\_lines(0, 98)} reads the whole table.\par
\texttt{notepad}: \good{Question Ledger} (sort 
by g\_2b, descending; limit 10) and \good{Evidence Table} (each value with its source line).\par
\texttt{bash}: \good{\texttt{rows.sort(key=lambda r: (-r[1], r[0]))}}\par
\good{\textit{``Wait, I need to re-check L11.}} \textit{[\dots] So whitewa01 has g\_2b = 0 and g\_3b = 9, g\_ss = 1.''}\par
Answer: all nine rows \good{(correct)}
\end{tcolorbox}\vspace{1pt}
\begin{tcolorbox}[casebox, title={\textbf{(b) Rewriting: Tailor removes the step name of the evolved harness but keeps the action it induced.}}]
\textbf{Before Tailor}\par
\textit{``\dots Let me start by doing a structural probe to understand the document structure. \harn{Step 0: Structural probe (preliminary, MANDATORY)}: call scan() to get a line-range map; call grep(pattern=`\^{}doc', use\_regex=true) to detect multi-document bundles.''}
\tcblower
\textbf{After Tailor}\par
\textit{``I need to find which document contains the string `KNGUKM'. \good{Let me start by scanning the document structure.}''}\par
The actions stay and are re-executed under the seed harness: \texttt{scan()}, \texttt{grep("KNGUKM")}, \texttt{grep("\^{}doc")}.
\end{tcolorbox}\vspace{1pt}
\begin{tcolorbox}[casebox, title={\textbf{(c) Without rewriting: the model cites a nonexistent mandate at test time and runs out of turns.}}]
\textbf{EvoIn w/o Tailor} \xmark\par
\textit{``\dots Let me start by scanning the document to understand its structure and content. First, let me perform the \harn{structural probe as mandated in step 0}.''}\par
\texttt{\harn{read\_file}(\dots)} $\to$ \textit{Tool `read\_file' not found}\par
After 29 tool calls: \harn{Maximum iteration limit reached}, no answer.
\tcblower
\textbf{EvoIn} \cmark\par
\textit{``\dots Let me start by scanning the document to understand its structure and content.''}\par
\good{Uses only seed-harness tools} (\texttt{scan}, \texttt{grep}, \texttt{read\_lines}).\par
After 13 tool calls: \good{``The correct answer is (D).''}
\end{tcolorbox}
\caption{\textbf{Cases} with trajectory excerpts. Red marks harness-specific or wrong content, and green marks what differs on the right. Appendix~\ref{app:cases} gives the full trajectories.}
\label{fig:cases}
\end{figure}

\textbf{Is harness evolution needed?}
To test whether the gains come from harness evolution, we fine-tune the target model on the same number of successful trajectories collected under the seed harness from the same categories (Seed-harness SFT in Table~\ref{tab:component_ablation}).
This control improves ID Pass by only 1.87 points and does not improve OOD Pass (18.36 vs.\ 18.99), whereas EvoIn improves them by 10.87 and 9.21 points; the OOD Score gain of the control comes from partial credit on MRCR.
Seed-harness data alone adds little, so the gains come from harness evolution.
Figure~\ref{fig:cases}(a) illustrates what the evolved harness adds on a Table QA question about the games each player of team WS4 played at second base in 1872: the evolved harness makes the base model record the requirements and the source line of each value, sort with bash, and recheck a doubtful row, whereas under the seed harness the model drops rows and misreads a column.
We also find an interesting behavior in some evolved harnesses, which first check the document length, read a short document in full before answering, and use search or chunked reading only for longer documents (Appendix~\ref{app:case_length}).

\textbf{Is rewriting needed?}
To test whether rewriting is needed, we train on the raw trajectories collected under the evolved harnesses for the same tasks, skipping Tailor (EvoIn w/o Tailor in Table~\ref{tab:component_ablation}).
ID performance changes little (Pass 45.00 vs.\ 46.83), but OOD Pass drops from 28.20 to 19.67, close to Base (18.99).
Raw trajectories often justify their steps with instructions and tools that exist only in the evolved harness, and the model repeats these references on unseen tasks where they do not apply.
Figure~\ref{fig:cases}(b) shows a typical edit, which drops a step name of the evolved harness but keeps the actions, and Figure~\ref{fig:cases}(c) shows what happens without it: on a LongBench v2 question, the model trained on raw trajectories attributes its first step to a structural probe ``as mandated in step 0'', calls a tool that the seed harness does not have, and runs out of turns without producing a final answer.
Across the six OOD benchmarks, this model exhausts the 30-turn budget of the seed harness on 19.4\% of examples, compared with 6.1\% for EvoIn.
Rewriting is therefore necessary for the internalized procedures to transfer to unseen tasks.

\textbf{Can the target model drive the pipeline by itself?}
The All-Qwen run checks whether EvoIn needs stronger external models by replacing the Claude proposer and the GLM-5.2 Tailor model with the target model itself.
It produces 6,123 demonstrations and still raises ID Pass from 35.96\% to 42.04\%, 56\% of the gain of the main system, but its OOD Pass stays at the Base level (18.90 vs.\ 18.99).
The target model can thus drive the pipeline by itself, although its gains are smaller than those of the main system and transfer little to unseen tasks.

\textbf{Does a stronger rollout model provide better demonstrations?}
Here the stronger GLM-5.3~\citep{glm5} replaces Qwen as the rollout model under the same best harnesses and Tailor pipeline, and we train the target model on its demonstrations or on Qwen demonstrations for the same 1,752 tasks (bottom block of Table~\ref{tab:component_ablation}).
GLM-5.3 demonstrations raise ID Pass from 46.52 to 52.43 and OOD Pass from 22.04 to 25.68, but OOD Score stays flat (49.65 vs.\ 49.36).
A stronger rollout model therefore provides better demonstrations, although part of the advantage does not transfer to unseen tasks.
The harnesses evolved with Qwen also work for a model from another family, so they are not tied to a single model.

\textbf{Does EvoIn work for another model family?}
Finally, we apply EvoIn to Gemma-4-31B-it~\citep{gemma4}, with Gemma as both the rollout target and the student, to see whether the method extends beyond Qwen.
The fine-tuned model improves ID Pass from 49.61\% to 53.04\% and OOD Pass from 49.08\% to 50.10\%.
EvoIn therefore also works for another model family.
Appendix~\ref{app:ablation_details} gives the implementation details of this run.





\section{Discussion}

\textbf{EvoIn vs. model-harness co-evolution.}
EvoIn and model-harness co-evolution pursue related but different objectives. 
Model-harness co-evolution aims to improve the coupled agent, allowing the model and harness to adapt to each other for better end-to-end performance. 
EvoIn instead focuses on internalizing reusable decision procedures discovered through harness evolution into the model itself. 
This helps separate improvements in the model's own reasoning from better model-harness compatibility.
It also encourages the learned procedures to remain useful beyond the particular harness in which they were discovered.
Moreover, we view EvoIn and model-harness co-evolution as complementary. 
EvoIn can serve as a useful approach for improving the model-side generality of co-evolution, rather than merely adapting the model to an evolved harness. 
Improvements such as new tools, workflows, or other runtime components cannot be fully absorbed into the model and should remain part of the evolved harness. 
Meanwhile, model-harness co-evolution helps the model better utilize the components in the harness.

\textbf{What else should be internalized?}
We also want to discuss what other agent capabilities could eventually be absorbed into the model. 
Several candidates are particularly natural. 
First, while tools themselves remain external, reusable knowledge about how to use them can be internalized. 
The model can learn when a tool is needed, which tool to choose, and how to reason over its outputs. 
Second, recurring experience can gradually become model priors rather than remain in an ever-growing external memory that must be repeatedly retrieved.
For example, an accountant may initially rely on notes for a recurring workflow, but after sufficient practice can complete the same process without consulting them. 
Finally, agents can internalize patterns in environmental observations, forming expectations about the consequences of their actions. 
For example, an experienced driver anticipates how the vehicle will respond before taking an action.
While EvoIn focuses on decision procedures, it provides a general recipe for this broader process by first expressing and validating capabilities externally and then transferring them into the model. 
A more ambitious direction is to let agents themselves learn which improvements should be internalized and which should remain external, and autonomously use EvoIn to internalize the parts.

\textbf{Limitations and future directions.}
Although EvoIn reduces the need for manual design, the current evolution process is still guided by a stronger agent rather than carried out entirely by the target agent itself. 
A natural next step is to let the target agent improve and internalize its own procedures, moving from agent-assisted improvement toward self-improvement.
Second, our experiments perform multiple rounds of procedure evolution but only one complete evolution-internalization cycle.
Once internalization produces a stronger model, the same process can be repeated to discover and learn procedures that were previously out of reach. 
Repeated EvoIn cycles provide a concrete path toward recursive self-improvement.
Third, while our evaluation covers dozens of benchmarks, many focus on context reasoning. 
We will extend EvoIn to different settings, such as software engineering, web interaction, and other long-horizon environments, to test its generality. 
Finally, as discussed above, decision procedures are only one class of capabilities that may benefit from internalization. 
We plan to extend EvoIn to other reusable capabilities that can later be absorbed into the model.

\textbf{Conclusion.}
In this paper, we propose EvoIn, a simple yet effective method for improving agents' decision-making by bridging harness evolution and model internalization.
Experiments show that EvoIn improves performance on diverse in-domain tasks, transfers to unseen benchmarks under the original harness, and also improves a model from another family.
Ablations and case studies further show that the gains require both harness evolution and rewriting, and that without rewriting the model cites instructions and tools that exist only in the evolved harness.
Combining EvoIn with model-harness co-evolution could enable agents to improve both what remains external and what can be internalized, leading to stronger and more general agents.

\section*{Acknowledgments}

We thank Xin Zhao and Pluto Zhou at Tencent for the helpful support and discussions. We also thank Jiayi Chen, Yujiong Shen, Ming Zhang, Xinyi Xu, and Chenhao Huang for their valuable discussions and feedback.







\bibliography{iclr2027_conference}
\bibliographystyle{iclr2027_conference}

\appendix
\section{Appendix}
\label{sec:appendix}

\subsection{ID Task Categories}
\label{app:id_categories}
Our training data cannot be released publicly, so Table~\ref{tab:app_id_categories} describes the task in each of the 23 ID categories and the task group it belongs to.
For each category, we reserve 100 examples for evolution and a disjoint 100 for evolution testing, giving 2,300 examples in each split.
Table~\ref{tab:app_id_results} reports the results for each category.

\begin{table}[ht]
  \centering
  \caption{\textbf{ID task categories}, grouped by the task groups in Table~\ref{tab:id_results}.}
  \label{tab:app_id_categories}
  \small
  \begin{tabular}{lp{0.56\linewidth}}
    \toprule
    Category & Task \\
    \midrule
    \rowcolor{groupgray}\multicolumn{2}{l}{\textit{Multi-document key retrieval}} \\
    Multi-Doc Key Lookup & Locate randomly generated keys in synthetic multi-document inputs, then answer the questions attached to them or combine their answers. \\
    Needle QA & Find the question that follows a given key in a specified document amid long distractor text, then answer it. \\
    Cross-Doc Key Aggregation & Evaluate keyed expressions across all documents and aggregate the results, e.g., by sum, maximum, or conditional operations. \\
    \rowcolor{groupgray}\multicolumn{2}{l}{\textit{Evidence-grounded QA}} \\
    Exam Reading Comprehension & Answer multiple-choice, translation, and short-answer questions on Chinese college-entrance-exam reading passages. \\
    Evidence-Located QA & Locate supporting evidence in real documents, such as papers, encyclopedia articles, and transcripts, then answer. \\
    Faithfulness Verification & Judge whether a document supports a given claim, or answer using only information from the document. \\
    Long-Document Extraction & Exhaustively extract or compare information in long documents, such as court rulings, reports, novels, and score tables. \\
    \rowcolor{groupgray}\multicolumn{2}{l}{\textit{Structured-data reasoning}} \\
    Table QA & Filter, sort, and compute over tables in documents and databases. \\
    Table Statistics & Answer statistical questions over Markdown tables. \\
    Multi-Step Structured Reasoning & Perform multi-step filtering and computation over XML tables, multiple tables, and code syntax trees. \\
    \rowcolor{groupgray}\multicolumn{2}{l}{\textit{Log and dialogue tracking}} \\
    Event-Log State Tracking & Derive the state of entities after a sequence of logged events from initial records and rules, with follow-up questions across turns. \\
    Group-Chat Counting & Answer counting, ranking, and set questions over group-chat logs, e.g., who sent the most messages. \\
    Structured Chat Analysis & Locate messages, count sender patterns, and identify social roles, e.g., who most often helps others, in JSON-formatted group chats. \\
    \rowcolor{groupgray}\multicolumn{2}{l}{\textit{In-context learning}} \\
    Many-Shot Classification & Classify a new input after many in-context examples whose labels are often arbitrary symbols. \\
    In-Context Translation & Translate low-resource languages using grammar notes and parallel examples given in the context. \\
    \rowcolor{groupgray}\multicolumn{2}{l}{\textit{Document-grounded generation}} \\
    Document Summarization & Summarize documents under length or format requirements. \\
    Document-Grounded Writing & Translate, rewrite, tabulate, or compose text based on a document. \\
    Open-Ended Document Requests & Handle open-ended user requests about documents, such as reviewing and ranking proposals or planning a presentation. \\
    \rowcolor{groupgray}\multicolumn{2}{l}{\textit{Complex instruction following}} \\
    Constrained Single-Turn Requests & Complete single-turn writing or organization requests with many explicit constraints. \\
    Multi-Turn Instruction Following & Follow instructions whose constraints are added or revised over multiple turns. \\
    Long-Source Deliverables & Produce deliverables such as tables, slides, or itineraries from long meeting transcripts or multiple sources. \\
    Context-Restricted Assistance & Answer multi-turn requests using only the provided material, as required by the system prompt. \\
    Agent Role Tasks & Act as a specified agent in a multi-agent system and complete its task in the required format. \\
    \bottomrule
  \end{tabular}
\end{table}

\begin{table}[ht]
  \centering
  \caption{\textbf{ID results by category} (percent).
  Models and $\Delta$ follow Table~\ref{tab:id_results}, each category has 100 evolution-test examples.}
  \label{tab:app_id_results}
  \small
  \setlength{\tabcolsep}{5pt}
  \begin{tabular}{lrrrrrr}
    \toprule
    & \multicolumn{3}{c}{Pass} & \multicolumn{3}{c}{Score} \\
    \cmidrule(lr){2-4}\cmidrule(lr){5-7}
    Category & Base & EvoIn & $\Delta$ & Base & EvoIn & $\Delta$ \\
    \midrule
    \rowcolor{groupgray}\multicolumn{7}{l}{\textit{Multi-document key retrieval}} \\
    Multi-Doc Key Lookup & 77.00 & 83.00 & $+$6.00 & 77.00 & 83.00 & $+$6.00 \\
    Needle QA & 61.00 & 55.00 & $-$6.00 & 61.00 & 55.00 & $-$6.00 \\
    Cross-Doc Key Aggregation & 33.00 & 52.00 & $+$19.00 & 33.00 & 52.00 & $+$19.00 \\
    \rowcolor{groupgray}\multicolumn{7}{l}{\textit{Evidence-grounded QA}} \\
    Exam Reading Comprehension & 62.00 & 73.00 & $+$11.00 & 62.00 & 73.00 & $+$11.00 \\
    Evidence-Located QA & 48.00 & 57.00 & $+$9.00 & 48.00 & 57.00 & $+$9.00 \\
    Faithfulness Verification & 48.00 & 59.00 & $+$11.00 & 48.00 & 59.00 & $+$11.00 \\
    Long-Document Extraction & 51.00 & 57.00 & $+$6.00 & 62.13 & 70.72 & $+$8.59 \\
    \rowcolor{groupgray}\multicolumn{7}{l}{\textit{Structured-data reasoning}} \\
    Table QA & 32.00 & 52.00 & $+$20.00 & 32.00 & 52.00 & $+$20.00 \\
    Table Statistics & 57.00 & 66.00 & $+$9.00 & 57.00 & 66.00 & $+$9.00 \\
    Multi-Step Structured Reasoning & 43.00 & 66.00 & $+$23.00 & 43.00 & 66.00 & $+$23.00 \\
    \rowcolor{groupgray}\multicolumn{7}{l}{\textit{Log and dialogue tracking}} \\
    Event-Log State Tracking & 68.00 & 77.00 & $+$9.00 & 74.42 & 82.33 & $+$7.91 \\
    Group-Chat Counting & 71.00 & 80.00 & $+$9.00 & 71.00 & 80.00 & $+$9.00 \\
    Structured Chat Analysis & 20.00 & 68.00 & $+$48.00 & 20.00 & 68.00 & $+$48.00 \\
    \rowcolor{groupgray}\multicolumn{7}{l}{\textit{In-context learning}} \\
    Many-Shot Classification & 32.00 & 48.00 & $+$16.00 & 32.00 & 48.00 & $+$16.00 \\
    In-Context Translation & 20.00 & 30.00 & $+$10.00 & 20.00 & 30.00 & $+$10.00 \\
    \rowcolor{groupgray}\multicolumn{7}{l}{\textit{Document-grounded generation}} \\
    Document Summarization & 38.00 & 54.00 & $+$16.00 & 38.00 & 54.00 & $+$16.00 \\
    Document-Grounded Writing & 9.00 & 21.00 & $+$12.00 & 9.00 & 21.00 & $+$12.00 \\
    Open-Ended Document Requests & 19.00 & 33.00 & $+$14.00 & 19.00 & 33.00 & $+$14.00 \\
    \rowcolor{groupgray}\multicolumn{7}{l}{\textit{Complex instruction following}} \\
    Constrained Single-Turn Requests & 8.00 & 10.00 & $+$2.00 & 54.38 & 65.92 & $+$11.54 \\
    Multi-Turn Instruction Following & 1.00 & 5.00 & $+$4.00 & 55.56 & 62.90 & $+$7.34 \\
    Long-Source Deliverables & 6.00 & 3.00 & $-$3.00 & 37.90 & 36.13 & $-$1.77 \\
    Context-Restricted Assistance & 7.00 & 8.00 & $+$1.00 & 35.30 & 37.82 & $+$2.52 \\
    Agent Role Tasks & 16.00 & 20.00 & $+$4.00 & 35.63 & 44.29 & $+$8.66 \\
    \midrule
    All 23 categories & 35.96 & 46.83 & $+$10.87 & 44.58 & 56.40 & $+$11.82 \\
    \bottomrule
  \end{tabular}
\end{table}

\subsection{OOD Benchmarks}
\label{app:ood_benchmarks}
The OOD test set contains 2,064 examples from six benchmarks, and Table~\ref{tab:ood_results} gives the number from each benchmark.
AA-LCR~\citep{artificialanalysis2025lcr} contains 100 questions that require reasoning across several real-world documents, mostly company documents, government consultations, industry reports, and legal and marketing texts.
BrowseComp-LongContext adapts BrowseComp~\citep{wei2025browsecomp} to a long-context setting, in which each question comes with more than a hundred web pages, most of which are distractors.
LongBench v2~\citep{bai2024longbench2} consists of 503 multiple-choice questions over long contexts, covering single- and multi-document QA, long in-context learning, dialogue histories, code repositories, and structured data.
MRCR~\citep{vodrahalli2024michelangelo} embeds many similar requests, such as several emails on the same topic, in a long multi-turn conversation and asks the model to reproduce a specified response, such as the sixth email about a given topic, beginning with a given random string.
Its official score is the string similarity between the response and the reference, and a response without the required prefix scores zero.
Oolong~\citep{bertsch2025oolong} presents many unlabeled short texts, such as SMS messages or questions, and asks aggregate questions such as which label is more frequent, so the model must label each text before counting or comparing; we use 300 examples from its validation split.
Table-Longer was constructed by us and cannot be released; the other five benchmarks are public.

\subsection{Harness Evolution and Tailor Details}
\label{app:pipeline_details}

\textbf{Seed harness.}
The generic workflow of the seed harness scans the input, searches for relevant evidence, reads the corresponding source lines, optionally uses \texttt{bash} for computation and \texttt{notepad} for intermediate state, and submits the answer through \texttt{complete\_task}.

\textbf{Candidate validation.}
A proposed harness may only add or replace whole files among the system prompt, the agent configuration, tool schemas, tool implementations, and middleware, and paths outside the harness directory are rejected.
Static validation compiles every tool implementation and middleware module, checks that each configured tool binds to an importable and callable symbol and that each middleware class can be instantiated with its configured arguments, and loads the full agent configuration.
Runtime validation then runs the harness end to end with the target model on one example from the evolution split; the run must finish without errors, but the answer need not be correct.
A harness that fails either check receives a score of zero and is not evaluated on the evolution-test split.

\textbf{Tailor rewriting.}
Tailor represents trajectories in the Agent Trajectory Interchange Format~\citep{li2026atif} and treats the system prompt, tool schemas, and call budget of the seed harness as the only target environment.
Before rewriting, it removes unsafe calls that provably do not affect the answer and replays the trajectory under the seed harness, so the rewriting model sees seed-harness observations.
The rewriting model also receives the reference answer, but only to check that its edits preserve the answer's meaning; its prompt forbids using the reference answer as evidence, and neither the repair call nor the checking call receives it.
All three calls use GLM-5.2 with thinking enabled and JSON-formatted output.
The rewriting model returns edits only for the steps that need them: it can rephrase the reasoning, notes, or final answer, remap a tool name or its arguments to a form the seed harness supports, mark a call for re-execution after its arguments change, or drop a failed, redundant, or unsupported step, and it declares a trajectory infeasible when it cannot be converted safely.
For instance, ``Step 0 assigns N=83 to the SHORT bucket (N $\le$ 120), so the mandatory route is \ldots'' becomes ``The document is short, so I can read it in full and then decide whether further search is needed,'' which keeps the decision to read a short document in full but drops the step name, the threshold, and the route name.
The edits also fix defects that make a trajectory unsuitable for fine-tuning, such as failed searches, reasoning that contradicts observations, wrong line citations, claims of full reading over truncated observations, and answers that rely on evidence not visible in the trajectory.
No call or observation is fabricated for a capability that the seed harness lacks, and replayed \texttt{bash} commands run under memory, CPU-time, file-size, and process-count limits with a timeout.

\textbf{Deterministic checks and repair.}
Among other conditions, the deterministic checks reject a trajectory whose reasoning retains wording from the evolved harness, claims that the input lacks some information without a preceding search or read, contradicts a search result, or quotes lines that do not appear in the replayed observations, as well as a final answer that violates the decimal precision required by the task.
They also reject trajectories with unreplayed or failed actions, repeated failed calls, or evidence gathered only through \texttt{scan} or \texttt{grep} without reading the source lines.
Repair targets only the failures found by the checks, is attempted at most once per trajectory, and is followed by another full replay and check.

\textbf{Iterative development.}
We developed the rewriting prompt, the checking prompt, and the deterministic checks in a closed loop on a fixed set of ten trajectories covering eight categories, documents of different lengths, and different tool chains, rerunning every version on the same trajectories.
Outputs of the first version passed all API, JSON, tool-schema, and final-answer checks, but a manual audit found only one of the ten directly usable for training, largely because harness templates removed from the reasoning remained in notepad arguments, truncated reads were treated as complete evidence, and failed searches and redundant steps were kept.
Subsequent rounds added seed-harness replay, evidence checks, and the independent checking model, and then tightened the checking prompt and the checks for visible evidence, contradictions, and residual templates, together with the handling of dropped steps, re-execution, and unsafe \texttt{bash} commands.
Because each round caught false positives that earlier rounds had admitted, the number of directly usable trajectories fell from five to three, one, and two over these rounds.
We then extended the decimal-precision requirement of a task from the first number in the final answer to every number in the final answer and in tool calls.
The largest improvement came from adding the targeted repair stage, after which eight of the ten trajectories were directly usable and the remaining two were explicitly flagged for re-sampling.
The frozen pipeline was then scaled to the full rollout pool with GLM-5.2 as both the rewriting model and the checking model, and \texttt{bash} commands were required to be actually replayed in the seed runtime.

\subsection{Proposal Prompt}
\label{app:proposal_prompt}
The proposal agent receives the system prompt below and, as its user message, an evolution query built from the analysis of the current harness's rollouts on the evolution split.
With three candidates per round, the system prompt assigns the three strategy axes shown, one to each candidate.
We reproduce both with typographic quotation marks and dashes written in ASCII, the working name of the project replaced by EvoIn, and the name of the internal agent framework replaced by a generic term; in the query template, angle brackets mark the fields filled in each round.

\begin{tcolorbox}[promptbox, title=Proposal System Prompt]
\begin{Verbatim}[breaklines]
You are the EvoIn Full-Harness Evolution Engine -- a meta-agent that improves a long-context agent harness from a structured evolution query. The query summarizes the sampled train tasks via failure-pattern clusters, passing-strategy exemplars, per-task stability notes, and the previous round's change attribution.

Your job is to propose candidate harness file changes that maximize the pass rate on this task class. The target agent answers hidden long-document questions through evidence tools.

Core rules:
1. Evidence-driven: every proposed change must be grounded in the failure clusters and representative tasks in the query.
2. General mechanism, not task hacks: never include a specific task answer, document content, or hard-coded keyword from one example as a rule.
3. Full-harness action space: you may edit systemprompt.md, context_agent.yaml, tools/*.tool.yaml, tool_impl/*.py, and middleware/*.py. The solving workflow described in systemprompt.md -- the ordering and entry conditions of its steps -- is itself part of this evolvable space, not a fixed template; you may reshape it when the evidence supports doing so, including adding a lightweight preliminary step that first inspects the given context (for example its length or structure) and lets the agent adapt how it gathers evidence accordingly. You may add new tools only if you also register them in context_agent.yaml, provide a tools/<name>.tool.yaml schema, and provide or reuse a callable binding. Do not create new tools lightly. A brand-new tool is allowed only under the strict last-resort conditions stated in the toolset_budget axis.
4. Component choice discipline: explain why the chosen component level is right. Tool misuse usually belongs in tool descriptions or system prompt. If stronger retrieval operators require tool_impl changes, handle them under the toolset_budget axis. Use middleware only as an execution-level fallback when prompt- and tool-level measures have repeatedly failed, also under the toolset_budget axis.
5. Preserve validated behavior: do not delete useful seed/current prompt rules unless the query gives evidence that they cause regressions. This does not freeze the workflow: you may reorder steps or add a lightweight preliminary check when the evidence suggests a single fixed sequence is not best for every task.
6. Runtime safety: do not modify external agent-framework source, data, judge, credentials, or absolute paths. All file paths must be relative to the harness root. Python must compile. Tool YAML schema and Python function signatures/argument handling must stay consistent.
7. Middleware integrity: if you edit LongToolOutputMiddleware, preserve the agent framework's Middleware interface and do not simplify away its original semantics unless the evidence explicitly justifies a targeted change.
8. Candidate orthogonality: each candidate must focus on its assigned strategy axis.

Produce exactly 3 candidate(s), on these axes:
- candidate 1 `retrieval_flow`: You MUST focus on the RETRIEVAL FLOW only: the order/granularity/coverage of scan -> grep -> read_lines (how to not miss evidence, how to broaden the search). Do NOT substantially change evidence-citation discipline, and do NOT add/remove tools.
- candidate 2 `evidence_discipline`: You MUST focus on EVIDENCE / CITATION DISCIPLINE only: every claim must carry a line number / verbatim quote, trust raw lines over summaries on conflict, and re-check evidence for each sub-item before finishing. Do NOT change the retrieval-flow structure or tools.
- candidate 3 `toolset_budget`: You MUST focus on the TOOL SUBSET / ITERATION BUDGET only: add or remove enabled tools (trim or complete the set) and tune max_iterations. You generally MODIFY existing components. Creating a brand-new tool is a last resort, allowed ONLY when (a) failure evidence shows the current tool set fundamentally cannot express the needed capability, AND (b) you state in why_this_change the added cost it imposes on downstream trajectory rewriting / SFT. The system prompt may only get the minimal wording needed to match the tool change, do NOT rewrite the solving strategy.

Each candidate must include a causal account for later attribution: failure_pattern, root_cause, predicted_fix, why_this_change, and component_choice.

Return ONLY a strict JSON array. No markdown fences, no prose, no comments. Each element:
{"name": str, "strategy_axis": str, "failure_pattern": str, "root_cause": str, "predicted_fix": str, "why_this_change": str, "component_choice": str, "system_prompt": str, "tools": [str], "max_iterations": int|null, "rationale": str, "file_changes": [{"path": str, "action": "replace|add", "content": str}]}

file_changes rules:
- Use complete file contents, not diffs.
- Allowed paths only: systemprompt.md, context_agent.yaml, tools/*.tool.yaml, tool_impl/*.py, middleware/*.py.
- If you change context_agent.yaml, keep valid YAML and preserve required complete_task as stop tool.
- If you add a tool, add all required files and make its binding importable from the harness root.
- Prefer small, targeted file changes; do not rewrite every component without evidence.
\end{Verbatim}
\end{tcolorbox}

\begin{tcolorbox}[promptbox, title=Evolution Query Template (User Message)]
\begin{Verbatim}[breaklines]
# EvoIn Evolution Query

## 1. Current Iteration Overview
- class, round, number of train tasks, train pass rate, pass/fail counts, stability counts

## 2. Current Harness
- current tools, available tools, mandatory tools, max_iterations

## 3. Failure Pattern Clusters
- <failure type> x<count> (<rate>): <general lesson>
  representatives: <task id>(<score>): <root cause>; ...

## 4. Passing Strategy Summary
- <task id>: <tool flow> | judge=<judge rationale>

## 5. Result Robustness
Judge each failure's robustness from its own trajectory (evidence strength + judge rationale). Do not make aggressive changes driven by fragile / incidental cases.
<tasks whose repeated samples disagree, if any>

## 6. Previous Iteration Change Attribution
<how the previous round's changes affected the pass rate>

## 7. Best-Ever Harness Summary
<best harness so far and its score>

## 8. Execution Instructions
Analyze failures -> group into pattern classes -> design general mechanisms -> propose 3 orthogonal candidate harnesses. Optimize the pass rate. Do not overfit to a specific task, document, or answer. Prefer changes that address large stable-fail clusters.
<high-leverage recommendations from the failure analysis, if any>

## 9. Current System Prompt
<current system prompt>

## Current Full-Harness Files
These are the editable harness files for this round. Use them when proposing file_changes.
<context_agent.yaml, tools/*.tool.yaml, tool_impl/*.py, and middleware/*.py, each truncated to a length cap>
\end{Verbatim}
\end{tcolorbox}

\subsection{Tailor Prompts}
\label{app:tailor_prompts}
The boxes below reproduce the system prompts of the rewriting, repair, and checking calls, with typographic quotation marks and dashes written in ASCII.
The user message of each call supplies the seed harness, the task, and the trajectory in the tagged fields that the prompt describes.

\begin{center}
\begin{tcolorbox}[promptbox, title=Rewriting Prompt]
\textbf{System prompt:}

\begin{Verbatim}[breaklines, fontsize=\footnotesize]
# Role and Objective

You are a **Harness Delabeling and SFT Quality Rewriter**. Rewrite a high-scoring trajectory produced under an evolved harness into an equivalent trajectory that:

1. could have been naturally produced under the seed harness; and
2. is a causally coherent, evidence-closed SFT demonstration.

Output a sparse edit list, never the full trajectory.

Training uses the seed system prompt, seed tools, and seed runtime budget. The evolved harness may have supplied extra procedures, tool guidance, schemas, capabilities, or budget. Preserve useful problem-solving behavior while removing evolved-only dependencies and high-confidence SFT defects.

**Minimally strip the packaging, preserve the valid solving skeleton, and reject what cannot be made evidence-closed without invention.**

# Inputs

The user message contains:

- `<seed_system_prompt>`: complete seed instructions.
- `<seed_tools>`: enabled seed tools, including guidance and argument schemas.
- `<runtime_budget>`: `{"seed_max_iterations": ...}`.
- `<document_profile>`: `{"num_lines": ...}`. Use it only to judge route plausibility; it is not answer evidence.
- `<task>`: `question` and `gold`. Gold is only for checking semantic preservation.
- `<preflight_sanitization>`: a deterministic local audit log of unsafe, provably non-load-bearing calls removed from the raw best rollout before seed replay. It is not evidence; do not reconstruct or rely on removed calls.
- `<best_trajectory>`: the sanitized, seed-replayed trajectory in **Agent Trajectory Interchange Format (ATIF-v1.5)**. It is a `{"steps": [...]}` JSON object in which each agent step contains its tool calls and matched observations (`tool_call_id` <-> `source_call_id`).

Inspect the complete trajectory, including observations. Only `source: "agent"` steps may be edited.

# Two Acceptance Gates

## Gate A: Seed reproducibility

For every retained reasoning span, message, tool call, tool argument, and final-answer span, ask:

> Given only the seed prompt, seed tools, seed budget, non-semantic routing metadata, and earlier visible observations, could the model plausibly have produced this content and action on its own?

## Gate B: SFT demonstration quality

The retained trajectory must also satisfy:

- every answer-bearing claim has a visible evidence path;
- reasoning is internally consistent with earlier steps;
- failed, malformed, or duplicate calls do not remain when they are non-load-bearing and safely removable;
- no false claim of complete coverage remains;
- no evolved-only template survives in any training-visible field;
- the final answer addresses the requested scope and uses a clean seed-compatible representation.

Honor explicit task-local numeric precision constraints across the entire `complete_task.result`, not only its first scalar. If the task permits at most two decimal places, do not retain a longer intermediate decimal in the final payload; use a supported rounded value or an exact expression instead.

Do not perform stylistic polishing. Repair only seed-compatibility defects and high-confidence SFT defects covered by these gates.

# Evidence Visibility Axiom

Judge evidence from the exact serialized ATIF content, not from what a tool may have returned before serialization.

- Text replaced by `[truncated ...]`, `...[truncated ...]...`, or any equivalent marker is **not visible evidence**.
- A requested line range or metadata such as `start`, `end`, and `total_lines` does not prove that omitted lines were visible.
- An observation with `is_error=true` is not evidence, even if it contains partial content.
- `scan` summaries and `grep` previews are navigation evidence, not raw-line support when the seed requires `read_lines`.
- A tool call without a matching successful observation proves no fact.
- Gold is never trajectory evidence.

Consequently, a trajectory must not say "I read the entire document" merely because it requested the entire range. The serialized observation itself must show complete, untruncated coverage.

# Citation and Entailment Integrity

Treat every retained `L<number>` reference, line range, quoted source phrase, and section-boundary claim as a factual assertion. Its quoted text must occur in the exact cited visible line or range, and its boundary must agree with the nearby visible headings and content. Repair or remove a wrong citation even if the answer itself is otherwise correct.

Direct paraphrase is allowed, but do not strengthen the source with new causal, evaluative, novelty, scope, or mechanism claims. Do not turn a weak proposition into a stronger one unless visible source supports it.

# Rewrite Principles

## 1. Recursively remove evolved-only packaging

Apply delabeling to every training-visible agent field:

- `reasoning_content`;
- `message`;
- tool names and arguments, including `notepad` content;
- final `complete_task.result`.

Remove unsupported pass names, procedure numbers, bucket names, mandatory thresholds, protocol citations, ledger schemas, and template headings. Preserve the underlying action when seed naturally supports it.

Listing a term in `ungrounded_terms` is not enough. Every harness-specific occurrence of that term or template must be edited, dropped, or made part of an explicit infeasibility decision.

## 2. Preserve seed-plausible adaptive routing

A trajectory may use document length, shape, or question type to choose a retrieval path. Do not normalize every trajectory into a fixed `scan -> grep -> read_lines` sequence.

Preserve a route decision when the document profile, task, seed tools, and visible observations make it plausible. Remove only unsupported labels and hard thresholds.

For a short-document full read:

- preserve it when the seed exposes `num_lines`, the call fits seed schema, and the matching observation is successful and untruncated;
- do not describe it as complete if the observation is truncated;
- if seed requires a prerequisite that is absent and repair requires inventing a call or observation, mark the trajectory infeasible.

## 3. Enforce visible evidence closure

Every retained final-answer fact and load-bearing intermediate conclusion must trace to:

- an earlier successful raw-line observation whose relevant text is actually visible; or
- a calculation that can be reproduced with seed tools from visible inputs.

When a broad `read_lines` call is truncated:

- if one existing call can be narrowed to the exact load-bearing range, use `tool_remap` on that call and add `reground_flag` for its `tool_call_id`;
- never retain the old observation as if it matched the remapped call;
- if one remapped call cannot recover all required evidence, set `feasible_under_seed=false`;
- never invent a missing narrow-read observation.

When several existing `read_lines` calls are available, you may remap each of those existing calls to a distinct necessary evidence window and reground all of them. Do not add a new call merely to fill a gap.

## 4. Remove high-confidence SFT defects

You may make minimal edits for:

- a failed or empty call caused by an obvious schema or query mistake;
- a duplicate call or reasoning span that adds no evidence;
- a local contradiction with earlier visible evidence;
- a wrong line reference, quotation-to-line mapping, or unsupported semantic amplification;
- a false coverage claim;
- a literal output-encoding artifact such as visible `\n` text where real line breaks are clearly intended;
- an answer-scope omission that is directly established by a visible outline or explicit user checklist.

Rules:

- Drop a failed or duplicate step only when it is non-load-bearing and later reasoning remains coherent; rephrase the next retained step if it refers to the dropped failure.
- Correct a contradiction only from already visible evidence. Otherwise mark the trajectory infeasible.
- Do not add missing answer facts merely because they appear in gold.
- If the final answer is materially incomplete or wrong and cannot be repaired from visible evidence without changing its meaning, set `feasible_under_seed=false`.

## 5. Preserve facts and answer meaning

Never alter a supported fact, quotation, line number, number, entity, or conclusion. If final presentation conflicts with the seed output contract, preserve the answer's meaning and change only its representation.

# Derive the Trajectory-to-Seed Difference

Before editing, produce all seven `derived_diff` fields:

1. **ungrounded_terms**: unsupported names, procedures, labels, protocol claims, templates, or numeric thresholds.
2. **tool_map**: renamed or reparameterized calls and their seed equivalents.
   - A same-name call is native only if its arguments validate against seed schema.
   - Set `behavior_equivalent=true` only when the mapped seed call would clearly return equivalent content.
3. **unsupported_tools**: capabilities with no seed-equivalent or reconstructible seed path.
4. **iteration_budget**: `seed_max_iterations` and `trajectory_agent_steps`.
5. **output_contract**: seed's answer-format requirement and whether the trajectory conflicts with it.
6. **evidence_provenance_gaps**: load-bearing claims unsupported by earlier visible raw lines or seed-available calculation. Each item contains `step_id`, `claim`, and `reason`.
7. **trajectory_quality_issues**: high-confidence SFT defects. Each item contains `step_id`, `type`, `detail`, and `resolution` (`edit | reground | infeasible`).

# Allowed Actions

Unlisted steps remain unchanged. Each edit uses one action:

1. **rephrase**: remove unsupported wording, repair a visibly grounded local contradiction, or normalize final presentation without changing supported facts.
   - To change `reasoning_content` or `message`, include the full replacement value in that field.
   - To change a seed-native tool argument without changing the tool itself, include that call's `tool_call_id` and an `arguments` object containing the replacement keys. The supplied keys overlay existing arguments; do not copy or edit observations.
   - Every `rephrase` must make at least one concrete replacement. A reason alone is not an edit.
2. **tool_remap**: map a renamed tool, schema mismatch, or overly broad existing read to seed-compatible name and arguments. Never invent an argument.
3. **reground_flag**: list affected `tool_call_id`s in `reground` when a remapped call must be re-executed. Never fabricate observations.
4. **drop**: remove a non-load-bearing unsupported, failed, empty, or duplicate step when coherence is preserved.
5. **unsupported_capability**: mark a load-bearing unavailable capability or evidence source and set `feasible_under_seed=false`.

Multiple edits may target the same step and are applied in listed order.

# Final Self-check

Before returning JSON, apply the edit list mentally to the full trajectory:

1. If a call is dropped or remapped, remove or rephrase every later reference that describes its old result.
2. Recheck every explicit line citation, quoted phrase, and boundary claim against the surviving replayed evidence.
3. Remove any answer language that is stronger than the visible source.
4. Set `sft_ready_after_edits=true` only when no unresolved issue remains in the final training-visible trajectory.

# Iteration Budget

If agent-step count exceeds `seed_max_iterations`:

1. Apply compatibility and evidence edits first.
2. Drop empty, failed, or clearly duplicate probes.
3. Preserve necessary raw evidence and causal coherence.
4. If the trajectory still cannot fit, set `feasible_under_seed=false`.

# Hard Constraints

- Never fabricate tool calls, observations, evidence, facts, or answer content.
- Never treat omitted or truncated text as visible.
- Never use `<task>.gold` as evidence or to reconstruct missing support.
- Never leave an identified evolved-only template in retained agent fields.
- Never claim full coverage after a truncated or failed read.
- Never force budget or quality compliance at the cost of correctness.
- Output exactly one valid JSON object with no code fence, comments, or surrounding text.

# Output Format

```json
{
  "derived_diff": {
    "ungrounded_terms": [],
    "tool_map": [
      {
        "traj_tool": "read_lines",
        "seed_tool": "read_lines",
        "behavior_equivalent": false
      }
    ],
    "unsupported_tools": [],
    "iteration_budget": {
      "seed_max_iterations": 30,
      "trajectory_agent_steps": 12
    },
    "output_contract": {
      "seed_requirement": "plain text or a numbered list",
      "final_answer_conflicts": false
    },
    "evidence_provenance_gaps": [],
    "trajectory_quality_issues": [
      {
        "step_id": 2,
        "type": "truncated_evidence",
        "detail": "the broad read omitted the answer-bearing lines from the serialized observation",
        "resolution": "reground"
      }
    ]
  },
  "feasible_under_seed": true,
  "sft_ready_after_edits": false,
  "infeasibility_reason": null,
  "overall_risk": "medium",
  "edits": [
    {
      "step_id": 2,
      "action": "tool_remap",
      "tool_call_id": "c2",
      "arguments": {
        "start": 48,
        "end": 55
      },
      "reason": "narrow the existing read to the load-bearing evidence range"
    },
    {
      "step_id": 2,
      "action": "reground_flag",
      "reground": ["c2"],
      "reason": "the remapped call must be executed against the seed tool"
    }
  ]
}
```

Field requirements:

- Always include all seven `derived_diff` fields; use empty arrays where appropriate.
- `overall_risk` is `none | light | medium | heavy`.
- `sft_ready_after_edits=true` only when all identified issues are resolved by edits, no regrounding remains, and the result is seed-reproducible and evidence-closed.
- `feasible_under_seed=true` with `sft_ready_after_edits=false` is valid only when a concrete unresolved issue remains, such as required regrounding.
- For a `rephrase` of tool arguments, always include `tool_call_id` and a non-empty `arguments` object.
- Do not emit a no-op edit. If no training-visible field changes, omit the edit.
- For a compact `tool_remap`, include `tool_call_id` when the step has multiple calls; include only changed `function_name` and/or `arguments`.
- A `reground_flag` edit includes a `reground` array.
- `reason` and `infeasibility_reason` are audit metadata and do not enter training data.

# Pre-Output Closure Check

1. Every retained action and argument is seed-plausible.
2. Every harness-specific term listed in `ungrounded_terms` has been removed from all retained agent fields, including tool arguments.
3. Same-name calls validate against seed schemas.
4. Every retained answer fact has an earlier successful, visible evidence path; truncated text and gold were not used.
5. No retained reasoning contradicts earlier visible evidence.
6. Failed, empty, and duplicate calls were either safely removed or explicitly justified.
7. No retained step falsely claims complete coverage.
8. The trajectory fits seed budget and ends with a valid closing step, or infeasibility is reported.
9. `sft_ready_after_edits` reflects unresolved regrounding and quality issues.
10. The output is one valid JSON object containing all required fields.
\end{Verbatim}
\end{tcolorbox}
\end{center}

\begin{center}
\begin{tcolorbox}[promptbox, title=Repair Prompt]
\textbf{System prompt:}

\begin{Verbatim}[breaklines, fontsize=\footnotesize]
# Role

You are a conservative **trajectory repair editor**. You receive a trajectory
that has already been rewritten and fully replayed under the seed harness,
together with structured deterministic-lint findings. Return only the smallest
sparse edit list needed to resolve the supplied blockers.

# Inputs and evidence boundary

The user message contains the seed system prompt, seed tools, runtime budget,
the original question, the current replayed trajectory, the Stage-2
`derived_diff`, and the initial and post-deterministic lint findings.

You never receive gold. Do not infer, request, reconstruct, or use gold. Only
earlier successful and non-truncated replayed observations are evidence.
Navigation output is not raw-line evidence when a raw read is required.

# Allowed repairs

Repair only these blocker families:

- `residual_template`;
- `unsupported_source_absence_claim`;
- `reasoning_observation_mismatch`;
- `reasoning_line_reference_mismatch`;
- `contradictory_document_key_mapping`;
- a demonstrably non-load-bearing `empty_navigation_call` or
  `repeated_failed_call`;
- `truncated_full_read_claim` when an existing call can be safely narrowed to
  a load-bearing visible range;
- `decimal_precision_violation`;
- `literal_newline_encoding`.

If a repair needs new evidence, a new tool call, a new observation, or facts
not visible in the trajectory, return `{"edits":[]}`. Never weaken a
load-bearing claim merely to hide a missing evidence path.

# Sparse-edit constraints

Only edit existing `source: "agent"` steps. Allowed actions are `rephrase`,
`tool_remap`, `reground_flag`, and `drop`, with the same sparse edit shape used
by the rewrite stage.

- Never add a tool call or change a `tool_call_id`.
- Never emit, copy, replace, or edit an `observation`.
- Never edit system or user steps.
- Preserve `complete_task.arguments.result` byte-for-byte unless the only
  difference is replacing literal `\n` with real line breaks or applying the
  question's explicit decimal-place requirement with ROUND_HALF_UP.
- A tool argument change must target an existing call. The pipeline will erase
  stale observations and replay the full retained trajectory.
- Drop a step only when it is provably non-load-bearing and no later retained
  text depends on its result.
- A broad truncated read may be remapped only by reusing an existing call. If
  the available calls cannot expose all load-bearing evidence, emit no repair.

# Output

Return exactly one JSON object and no surrounding text:

```json
{
  "edits": [
    {
      "step_id": 4,
      "action": "rephrase",
      "reasoning_content": "The prior grep returned the cited match.",
      "reason": "remove a contradiction with the replayed observation"
    }
  ]
}
```

Every edit must make a concrete change. Do not include analysis, full
trajectories, lint findings, observations, or proposed new calls in the output.
\end{Verbatim}
\end{tcolorbox}
\end{center}

\begin{center}
\begin{tcolorbox}[promptbox, title=Checking Prompt]
\textbf{System prompt:}

\begin{Verbatim}[breaklines, fontsize=\footnotesize]
# Role

You are an independent **SFT admission judge**. You do not rewrite trajectories.
Decide whether one proposed seed-harness trajectory is a high-quality supervised
fine-tuning demonstration.

The trajectory has already passed structural validation, seed-tool replay, and
deterministic linting. Re-check their conclusion rather than assuming it is
correct. Be conservative: acceptance means this example is safe to train on,
not merely that it looks plausible.

# Inputs

The user message contains:

- `<seed_system_prompt>`, `<seed_tools>`, and `<runtime_budget>`;
- `<task>` with the original question only. The judge never receives gold;
- `<deterministic_gate>` with replay and lint findings;
- `<rewrite_audit>` with the Stage-2 self-assessment, derived issues, and edits actually applied;
- `<rewritten_trajectory>` with all training-visible steps and replayed observations.

# Admission Criteria

Accept only if all conditions hold:

1. The retained workflow and every tool argument are naturally plausible under
   the seed harness.
2. The trajectory has causal coherence: later reasoning and the final answer
   follow from earlier successful, visible observations. Read every retained
   `reasoning_content`, message, scratchpad/notepad value, and final answer;
   do not judge from the final answer alone.
3. Every answer-bearing factual claim has adequate visible support. Truncated,
   errored, missing, navigation-only, or stale observations are not support.
4. No evolved-harness labels, protocol templates, fabricated evidence, false
   coverage claims, literal output-encoding artifacts, failed/empty
   non-load-bearing calls, or contradictory behavior remain.
5. The final answer answers the requested scope without relying on gold. When
   the visible source outline or question explicitly enumerates required
   aspects, reject a material omission rather than accepting a merely
   factually correct partial answer.
6. The example is pedagogically useful: it demonstrates a compact, legible
   evidence-gathering path rather than preserving redundant or contradictory
   behavior.

Every explicit `L<number>` citation, line range, quoted source phrase, and
section-boundary claim must match the exact preceding replayed lines. Reject
unsupported semantic amplification and any decimal that violates an explicit
numeric precision instruction.

Reject when a load-bearing assertion cannot be checked from the serialized
trajectory. Do not reject merely for natural language style differences or
because a short context is read directly when that is seed-plausible.

A later correction does not erase a false retained claim. Reject if any
training-visible reasoning wrongly says evidence is absent, assigns an entity,
document, key, line, option, or quantity inconsistently, or otherwise
contradicts the replayed observations--even when the final answer happens to be
correct.

# Mandatory Review Passes

Before deciding, perform all four checks:

1. **Trace pass:** compare every retained reasoning claim and final-answer
   claim against the preceding replayed observations.
2. **Scope pass:** compare the answer against the exact user request and any
   visible section outline/checklist.
3. **Edit-integrity pass:** inspect `<rewrite_audit>` together with the final
   trajectory. If an intended repair is absent from the training-visible
   result, or a residual template/failed probe remains, reject it.
4. **Citation-and-entailment pass:** check every retained line citation and
   quoted phrase against the exact replayed line(s), then check that the final
   answer did not strengthen the source through unsupported interpretation.

Do not trust a statement in `<rewrite_audit>` that a repair occurred. Verify
the final serialized trajectory itself.

# Output

Return exactly one valid JSON object, with no markdown fence or extra text:

```json
{
  "accept": true,
  "quality_risk": "none",
  "blockers": [],
  "warnings": [],
  "reviewed_agent_step_ids": [1, 2, 3],
  "consistency_findings": [],
  "rationale": "The final answer is supported by replayed raw-line evidence and the retained workflow is seed-plausible."
}
```

- `quality_risk` must be one of `none | light | medium | heavy`.
- Each blocker or warning is an object with `type`, optional `step_id`, and
  `detail`.
- `reviewed_agent_step_ids` must list every agent `step_id` in the supplied
  rewritten trajectory exactly once.
- `consistency_findings` is an array of unresolved false claims or
  cross-step contradictions. `accept=true` requires it to be empty.
- `accept=true` requires empty `blockers`, `warnings`, and
  `consistency_findings` arrays. Any known quality caveat means the example is
  not yet SFT-ready.
- Do not suggest edits, invent missing evidence, or use gold to repair the
  trajectory.
\end{Verbatim}
\end{tcolorbox}
\end{center}

\subsection{SFT Configuration}
\label{app:sft}
After removing duplicates and test overlaps and dropping demonstrations longer than 24,576 tokens, we obtain 12,035 SFT demonstrations.
Table~\ref{tab:app_sft} lists the SFT configuration shared by the main model and all Qwen ablations.

\begin{table}[h]
  \centering
  \caption{\textbf{SFT configuration} for the main model and all Qwen ablations.}
  \label{tab:app_sft}
  \small
  \begin{tabular}{ll}
    \toprule
    Setting & Value \\
    \midrule
    Fine-tuning & Full-parameter SFT \\
    Framework & Megatron-Bridge~\citep{shoeybi2019megatron} \\
    Epochs & 3 \\
    Sequence length & 24,576 tokens \\
    Global / micro batch size & 8 / 1 \\
    Tensor / expert parallelism & 8 / 8 \\
    Precision & bfloat16 \\
    Optimizer & Adam, $(\beta_1,\beta_2)=(0.9,0.95)$, $\epsilon=10^{-8}$ \\
    Weight decay & 0.1 \\
    Gradient clipping & 1.0 \\
    Learning rate & Cosine decay from $5\times10^{-6}$ to $5\times10^{-7}$ \\
    Warmup & First 3\% of optimizer steps \\
    Optimizer steps (main model) & 1,488 per epoch, 4,464 in total \\
    Training seed & 42 \\
    \bottomrule
  \end{tabular}
\end{table}

\subsection{Evaluation Settings}
\label{app:eval}
Each ID category is scored by either a Qwen3.5-35B-A3B judge or a GPT-OSS-120B judge~\citep{openai2025gptoss}.
LongBench v2, MRCR, and Oolong are scored locally with their official scoring rules, and the other three OOD benchmarks use GPT-OSS-120B as the judge.
Table~\ref{tab:app_eval} lists the decoding settings of the evaluated Qwen models and of the two LLM judges.
We use fixed denominators: a missing target or judge score counts as zero rather than being dropped.
The directly compared OOD runs have no target or judge errors.

\begin{table}[h]
  \centering
  \caption{\textbf{Decoding settings in the Qwen evaluations.}
  The GPT-OSS-120B column lists the judge settings for the OOD benchmarks.
  A dash marks a setting that is not set explicitly.}
  \label{tab:app_eval}
  \small
  \begin{tabular}{lccc}
    \toprule
    Setting & Evaluated model & Qwen3.5-35B-A3B judge & GPT-OSS-120B judge \\
    \midrule
    Context limit (tokens) & 180,224 & -- & -- \\
    Max generated tokens & 16,000 & 8,192 & 8,192 \\
    Temperature & 1.0 & 0 & 0.7 \\
    Top-$p$ & 1.0 & 1.0 & 0.95 \\
    Top-$k$ & 50 & -- & 50 \\
    Repetition penalty & 1.05 & -- & 1.05 \\
    Thinking & Enabled & Disabled & Default \\
    Seed & 1234 per task & -- & -- \\
    \bottomrule
  \end{tabular}
\end{table}

\subsection{Ablation Details}
\label{app:ablation_details}
Unless stated otherwise, the runs in Section~\ref{sec:further_analysis} follow the setup in Section~\ref{sec:experiment_setup} and report epoch-3 results.
Seed-harness SFT samples successful seed-harness trajectories deterministically, stratified by data source and category, to match the corpus size of EvoIn.
EvoIn w/o Tailor keeps the task, system prompt, tools, and split of each example whenever possible; 37 of the 12,035 examples (0.31\%) are replaced by another example from the same stratum because the raw trajectory is too long or does not end with a valid final answer.
The rollout comparison uses 1,752 task-paired examples from 22 of the 23 ID categories, and its Qwen arm reuses the corresponding demonstrations of the main corpus.
The Gemma run produces 15,008 training and 138 validation demonstrations.
It uses its own tool parser, training stack, decoding policy, and OOD judge version, so its results are not directly comparable with those of the Qwen runs.
Tables~\ref{tab:app_ablation_id} and~\ref{tab:app_ablation_ood} report per-category ID results and per-benchmark OOD results for every run in Section~\ref{sec:further_analysis}, alongside Base and EvoIn.

\begin{table}[ht]
  \centering
  \caption{\textbf{Per-category ID Score of all runs} (percent).
  The ablation columns correspond to the questions in Section~\ref{sec:further_analysis}: Seed-harness SFT removes harness evolution, EvoIn w/o Tailor removes rewriting, All-Qwen uses the target model as both the proposer and the Tailor model, the two rollout columns are the task-paired Qwen and GLM-5.3 arms, and the Gemma columns repeat the pipeline with Gemma-4-31B-it.}
  \label{tab:app_ablation_id}
  \footnotesize
  \setlength{\tabcolsep}{2pt}
  \begin{tabular}{lrrrrrrrrr}
    \toprule
    & \multicolumn{7}{c}{Qwen3.5-35B-A3B} & \multicolumn{2}{c}{Gemma} \\
    \cmidrule(lr){2-8}\cmidrule(lr){9-10}
    Category & Base & EvoIn & \makecell{Seed-\\harness} & \makecell{w/o\\Tailor} & \makecell{All-\\Qwen} & \makecell{Qwen\\rollout} & \makecell{GLM-5.3\\rollout} & Base & EvoIn \\
    \midrule
    \rowcolor{groupgray}\multicolumn{10}{l}{\textit{Multi-document key retrieval}} \\
    Multi-Doc Key Lookup & 77.00 & 83.00 & 74.00 & 86.00 & 84.00 & 87.00 & 89.00 & 75.00 & 86.00 \\
    Needle QA & 61.00 & 55.00 & 62.00 & 62.00 & 66.00 & 73.00 & 79.00 & 59.00 & 76.00 \\
    Cross-Doc Key Aggregation & 33.00 & 52.00 & 35.00 & 55.00 & 43.00 & 47.00 & 70.00 & 46.00 & 53.00 \\
    \rowcolor{groupgray}\multicolumn{10}{l}{\textit{Evidence-grounded QA}} \\
    Exam Reading Comprehension & 62.00 & 73.00 & 60.00 & 76.00 & 60.00 & 69.00 & 82.00 & 72.00 & 75.00 \\
    Evidence-Located QA & 48.00 & 57.00 & 50.00 & 54.00 & 56.00 & 58.00 & 75.00 & 55.00 & 59.00 \\
    Faithfulness Verification & 48.00 & 59.00 & 56.00 & 54.00 & 55.00 & 58.00 & 60.00 & 61.00 & 70.00 \\
    Long-Document Extraction & 62.13 & 70.72 & 71.06 & 69.99 & 70.49 & 73.00 & 78.38 & 78.05 & 80.30 \\
    \rowcolor{groupgray}\multicolumn{10}{l}{\textit{Structured-data reasoning}} \\
    Table QA & 32.00 & 52.00 & 44.00 & 60.00 & 57.00 & 51.00 & 44.00 & 51.00 & 62.00 \\
    Table Statistics & 57.00 & 66.00 & 63.00 & 68.00 & 72.00 & 69.00 & 74.00 & 78.00 & 78.00 \\
    Multi-Step Structured Reasoning & 43.00 & 66.00 & 47.00 & 66.00 & 52.00 & 65.00 & 58.00 & 66.00 & 75.00 \\
    \rowcolor{groupgray}\multicolumn{10}{l}{\textit{Log and dialogue tracking}} \\
    Event-Log State Tracking & 74.42 & 82.33 & 78.00 & 80.33 & 81.58 & 85.50 & 85.50 & 82.00 & 83.50 \\
    Group-Chat Counting & 71.00 & 80.00 & 74.00 & 74.00 & 70.00 & 79.00 & 78.00 & 81.00 & 82.00 \\
    Structured Chat Analysis & 20.00 & 68.00 & 30.00 & 58.00 & 52.00 & 64.00 & 67.00 & 71.00 & 67.00 \\
    \rowcolor{groupgray}\multicolumn{10}{l}{\textit{In-context learning}} \\
    Many-Shot Classification & 32.00 & 48.00 & 26.00 & 37.00 & 27.00 & 41.00 & 37.00 & 72.00 & 56.00 \\
    In-Context Translation & 20.00 & 30.00 & 16.00 & 21.00 & 22.00 & 26.00 & 18.00 & 28.00 & 28.00 \\
    \rowcolor{groupgray}\multicolumn{10}{l}{\textit{Document-grounded generation}} \\
    Document Summarization & 38.00 & 54.00 & 37.00 & 45.00 & 52.00 & 53.00 & 70.00 & 60.00 & 64.00 \\
    Document-Grounded Writing & 9.00 & 21.00 & 13.00 & 20.00 & 12.00 & 17.00 & 31.00 & 25.00 & 25.00 \\
    Open-Ended Document Requests & 19.00 & 33.00 & 23.00 & 22.00 & 24.00 & 33.00 & 58.00 & 30.00 & 44.00 \\
    \rowcolor{groupgray}\multicolumn{10}{l}{\textit{Complex instruction following}} \\
    Constrained Single-Turn Requests & 54.38 & 65.92 & 57.45 & 69.84 & 61.86 & 63.42 & 63.11 & 70.48 & 78.90 \\
    Multi-Turn Instruction Following & 55.56 & 62.90 & 58.94 & 59.03 & 58.09 & 61.80 & 58.94 & 72.10 & 72.74 \\
    Long-Source Deliverables & 37.90 & 36.13 & 32.39 & 37.23 & 34.58 & 37.65 & 40.64 & 43.39 & 44.49 \\
    Context-Restricted Assistance & 35.30 & 37.82 & 37.22 & 41.74 & 38.01 & 42.02 & 39.96 & 47.03 & 44.33 \\
    Agent Role Tasks & 35.63 & 44.29 & 41.61 & 46.31 & 43.64 & 41.46 & 52.08 & 34.71 & 36.70 \\
    \midrule
    All 23 categories & 44.58 & 56.40 & 47.25 & 54.89 & 51.84 & 56.30 & 61.24 & 59.03 & 62.65 \\
    \bottomrule
  \end{tabular}
\end{table}

\begin{table}[ht]
  \centering
  \caption{\textbf{Per-benchmark OOD results of all runs} (percent).
  Columns follow Table~\ref{tab:app_ablation_id}.
  The Qwen-rollout and Gemma columns use the same GPT-OSS-120B judge served from a different endpoint, and Gemma also uses its own evaluation protocol.}
  \label{tab:app_ablation_ood}
  \footnotesize
  \setlength{\tabcolsep}{2pt}
  \begin{tabular}{lrrrrrrrrrr}
    \toprule
    & & \multicolumn{7}{c}{Qwen3.5-35B-A3B} & \multicolumn{2}{c}{Gemma} \\
    \cmidrule(lr){3-9}\cmidrule(lr){10-11}
    Benchmark & $N$ & Base & EvoIn & \makecell{Seed-\\harness} & \makecell{w/o\\Tailor} & \makecell{All-\\Qwen} & \makecell{Qwen\\rollout} & \makecell{GLM-5.3\\rollout} & Base & EvoIn \\
    \midrule
    \rowcolor{groupgray}\multicolumn{11}{l}{\textit{Pass}} \\
    AA-LCR & 100 & 27.00 & 37.00 & 30.00 & 36.00 & 34.00 & 37.00 & 49.00 & 58.00 & 64.00 \\
    BrowseComp-LongContext & 295 & 12.54 & 13.56 & 11.19 & 12.20 & 15.25 & 12.88 & 19.32 & 22.03 & 32.88 \\
    LongBench v2 & 503 & 33.20 & 44.73 & 25.25 & 21.87 & 36.58 & 36.98 & 36.78 & 58.05 & 61.43 \\
    MRCR & 800 & 2.00 & 8.88 & 6.88 & 9.50 & 3.00 & 5.75 & 15.00 & 50.38 & 43.12 \\
    Oolong & 300 & 43.33 & 60.33 & 39.33 & 42.33 & 29.33 & 43.33 & 36.33 & 55.67 & 60.67 \\
    Table-Longer & 66 & 22.73 & 42.42 & 24.24 & 31.82 & 22.73 & 27.27 & 15.15 & 42.42 & 56.06 \\
    Weighted average & 2,064 & 18.99 & 28.20 & 18.36 & 19.67 & 18.90 & 22.04 & 25.68 & 49.08 & 50.10 \\
    \rowcolor{groupgray}\multicolumn{11}{l}{\textit{Score}} \\
    AA-LCR & 100 & 27.00 & 37.00 & 30.00 & 36.00 & 34.00 & 37.00 & 49.00 & 58.00 & 64.00 \\
    BrowseComp-LongContext & 295 & 12.54 & 13.56 & 11.19 & 12.20 & 15.25 & 12.88 & 19.32 & 22.03 & 32.88 \\
    LongBench v2 & 503 & 33.20 & 44.73 & 25.25 & 21.87 & 36.58 & 36.98 & 36.78 & 58.05 & 61.43 \\
    MRCR & 800 & 32.55 & 76.31 & 70.14 & 75.95 & 63.34 & 75.54 & 76.66 & 91.83 & 92.30 \\
    Oolong & 300 & 44.58 & 61.42 & 41.26 & 44.22 & 30.18 & 45.12 & 36.80 & 58.53 & 63.77 \\
    Table-Longer & 66 & 22.73 & 42.42 & 24.24 & 31.82 & 22.73 & 27.27 & 15.15 & 42.42 & 56.06 \\
    Weighted average & 2,064 & 31.02 & 54.49 & 43.16 & 45.70 & 42.41 & 49.36 & 49.65 & 65.56 & 69.61 \\
    \bottomrule
  \end{tabular}
\end{table}

\subsection{Case Studies}
\label{app:cases}
This section gives the three cases of Figure~\ref{fig:cases} in more detail and adds a case in which the evolved harness chooses its procedure by document length.
The first case shows harness evolution refining steps that the seed harness already has, such as keeping notes, computing with \texttt{bash}, and checking the answer before submission.
The last case shows a new behavior that harness evolution adds through candidates that raise the evolution-test Pass.
In 11 of the 23 categories, the best harness first checks the document length, reads a short document in full before answering, and uses search or chunked reading only for longer documents.
Quotations are verbatim except for elisions marked ``[...]'', and tool observations are summarized after \texttt{->}.

\textbf{Harness evolution (Figure~\ref{fig:cases}(a)).}
The case comes from the Table QA category, where the base model solves 34 of the 100 evolution-test questions under the seed harness and 61 under the evolved harness.
Of the 32 questions that only the evolved harness solves, about 25 fail under the seed harness only because the answer includes reasoning or numbering; the case below is one of the remaining content errors.
Among other changes, the evolved harness adds the following workflow rules to the system prompt.

\newpage

\begin{tcolorbox}[promptbox, title=Evolved Table QA Harness (Excerpt)]
\begin{Verbatim}[breaklines, fontsize=\scriptsize]
1. Read the question. Write the Question Ledger to notepad: required headers verbatim (...), format directives, target N, required answer shape.
4. read_lines() over a wide range covering the full relevant row block, INCLUDING the header row and separator row of the source table.
5. Build an Evidence Table in notepad. For every header AND every cell you plan to emit in result, record a row: <emitted_token> | Lk | <verbatim source token>
6. bash() (python3) for any sort, aggregation, arithmetic. Sorting uses an explicit primary key AND an explicit tie-breaker.
7. Sub-item recheck (mandatory): iterate the Evidence Table before submission and confirm each row's emitted token equals its source token.
\end{Verbatim}
\end{tcolorbox}

The question asks: ``Find the player IDs and number of games played at 2nd base for players on team WS4 in 1872, ordered by the number of 2nd base games from highest to lowest, and limit the results to 10.''
The reference answer lists nine players, holliho01 with 9 games and eight players with 0 games.

\begin{tcolorbox}[promptbox, title=Seed Harness (Fails)]
\begin{Verbatim}[breaklines, fontsize=\scriptsize]
[A1] scan() -> one 98-line table
[A2] grep(pattern="WS4", useRegex=false) -> TypeError (wrong parameter name)
[A3] grep(pattern="WS4", use_regex=false, context_lines=5) -> matches at L4-L12
[A4] read_lines(4, 13) -> the nine WS4 rows
[A5] "[...] L11: whitewa01, g_2b = 0 [...] For players with zero games, I need to decide whether to include them. [...] I'll return only those who actually played at 2nd base, ordered by games from highest to lowest."
     read_lines(3, 4) -> the header row
[A6] "[...] Filtering for non-zero 2nd base games, only two players qualify: holliho01 and whitewa01, each with 9 games. [...]"
     complete_task(result="holliho01\t9\nwhitewa01\t1")
\end{Verbatim}
\end{tcolorbox}

\begin{tcolorbox}[promptbox, title=Evolved Harness (Passes)]
\begin{Verbatim}[breaklines, fontsize=\scriptsize]
[A1] scan() -> one 98-line table
[A2] read_lines(0, 98) -> the whole table
[A3] notepad(action="write", content="Question Ledger: [...] - Sort: by g_2b descending (highest to lowest) - Limit: 10 rows - Output format: only the answer, no explanation / Evidence Table (WS4 1872 players): | player_id | g_2b | Lk | | holliho01 | 9 | L4 | [...] | whitewa01 | 0 | L11 | | yeatmbi01 | 0 | L12 | [...]")
[A4] bash(command="python3 [...] rows.sort(key=lambda r: (-r[1], r[0])) [...]") -> holliho01,9 lennobi01,0 [...] yeatmbi01,0
[A5] "[...] Wait, I need to re-check L11. [...] So whitewa01 has g_2b = 0 and g_3b = 9, g_ss = 1. [...]"
     complete_task(result=<the nine rows as a table>)
\end{Verbatim}
\end{tcolorbox}

\textbf{Rewriting (Figure~\ref{fig:cases}(b)).}
The trajectory comes from the Multi-Doc Key Lookup category and is one of the 5,051 accepted trajectories in which Tailor removes references specific to the evolved harness.
The task asks which of three documents contains the key KNGUKM.
Tailor edits three of the five steps, and its checking model accepts the result without blockers.

\newpage

\begin{tcolorbox}[promptbox, title=Tailor Edits]
\begin{Verbatim}[breaklines, fontsize=\scriptsize]
Step 1, rephrase ("remove evolved-only template language ('Step 0', 'Structural probe', 'MANDATORY', 'preliminary')")
  before: "I need to find which document contains the key "KNGUKM". Let me start by doing a structural probe to understand the document structure. Step 0: Structural probe (preliminary, MANDATORY) - Call scan() to get a line-range map - Call grep(pattern='^doc', use_regex=true) to detect multi-document bundles. Let me start with these."
  after:  "I need to find which document contains the string "KNGUKM". Let me start by scanning the document structure."
  action kept: scan()
Steps 2-3, unchanged
  grep(pattern="KNGUKM") -> a match at L90
  grep(pattern="^doc", use_regex=true) -> document boundaries at L1, L64, and L83
Step 4, tool_remap and reground_flag ("remap non-load-bearing notepad call to read_lines confirming doc3 boundary (L83) and KNGUKM line (L90)")
  before: notepad(action="write", content="doc_boundaries: [...] target_line: L90 [...] subquestions: [...] format_spec: [...] coverage_checklist: [...]")
  after:  read_lines(83, 91), re-executed under the seed harness -> L83 starts doc3 and L90 contains KNGUKM
Step 5, rephrase ("remove evolved term 'structural probe' and add read_lines confirmation reference")
  the reasoning cites the new read_lines evidence; the final answer keeps doc3 and drops an internal tool name and a long quotation.
\end{Verbatim}
\end{tcolorbox}

\textbf{Without rewriting (Figure~\ref{fig:cases}(c)).}
Among the 321 OOD questions that EvoIn solves and EvoIn w/o Tailor fails, the failing trajectory of 76 contains terms from the evolved harnesses or calls a tool that the seed harness lacks; the case below is one of the 36 such questions from LongBench v2.
The question asks: ``Which of the following suggestions for the new year's initiatives is correct? (A) Ensure the looseness and progressiveness of policies, and strengthen the coordination and cooperation of various policies. (B) Support and highlight the dominant position of enterprises in production, and increase the cultivation of technical production talents. (C) Promote the growth of the RMB exchange rate at a reasonable and balanced level. (D) On the basis of a comprehensive and scientific summary of epidemic prevention and control work, we should do a more scientific, accurate, and efficient job in prevention and control.''
The reference answer is (D).

\begin{tcolorbox}[promptbox, title=EvoIn w/o Tailor (Fails)]
\begin{Verbatim}[breaklines, fontsize=\scriptsize]
[A1]  "I need to find information about new year's initiatives and which suggestion is correct among the given choices. Let me start by scanning the document to understand its structure and content. First, let me perform the structural probe as mandated in step 0."
      scan()
[A2-A7] grep and read_lines calls
[A8]  "I need to find the specific suggestions for new year's initiatives in this government work report. [...]"
      read_file(path="/tmp/[...]/read_lines_[...].txt") -> Error: Tool 'read_file' not found
[A9-A29] read_lines and grep calls
[A30] [Error: Maximum iteration limit reached.] -> no answer is submitted
\end{Verbatim}
\end{tcolorbox}

\begin{tcolorbox}[promptbox, title=EvoIn (Passes)]
\begin{Verbatim}[breaklines, fontsize=\scriptsize]
[A1]  "I need to find information about new year's initiatives and which suggestion is correct. Let me start by scanning the document to understand its structure and content."
      scan(chunk_size=100, max_chunks=20)
[A2-A12] grep and read_lines calls
[A13] complete_task(result="The correct answer is (D)")
\end{Verbatim}
\end{tcolorbox}

\textbf{Length check before retrieval.}
\label{app:case_length}
The proposal prompt allows the evolved workflow to begin with a lightweight check of the input (Appendix~\ref{app:proposal_prompt}), but it does not specify the check or the steps that follow it.
In each of the 11 categories with such a rule, the length check first enters the harness with the candidate selected in its round, which raises the evolution-test Pass over its parent by 1 to 17 points.
The thresholds differ across categories; for example, Faithfulness Verification treats a document of at most about 120 lines as short, and Evidence-Located QA one of at most 300 lines.
In Faithfulness Verification, a length check first appears in round 4 and is refined into the three size classes below in round 7, whose harness is selected as the best.

\begin{tcolorbox}[promptbox, title=Evolved Faithfulness Verification Harness (Excerpt)]
\begin{Verbatim}[breaklines, fontsize=\scriptsize]
### Step 0 - Preliminary shape check (MANDATORY FIRST STEP)
Always call `scan()` first. From the scan result note:
- Total lines N.
- Whether the doc is short (N <= ~120), medium (120 < N <= ~600), or long (N > 600).
- Salient section boundaries.
Then silently classify the task: verify | compare | fill-in | enumerate | dimension/attribute | open.

Based on (task_class, size) pick a retrieval strategy:
- SHORT doc: replace grep with a single full-doc `read_lines(0, -1)` sweep, then optionally grep for refinement.
- MEDIUM doc, targeted task: grep with multiple query variants, then `read_lines` around each hit with +/-7 lines.
- LONG doc, targeted task: grep multi-variant + `read_lines` around each hit.
- ANY size, enumerate / 'find all' / dedup / list task: BOTH multi-variant grep AND a chunked `read_lines` sweep of the entire document (chunks of ~200 lines) - do not rely on grep alone.
\end{Verbatim}
\end{tcolorbox}

Under this harness, the model states the size class after \texttt{scan} and follows the corresponding route, as the two tasks below show.
In the first, the document is a 36-line comment on a physics paper, and the question embedded in it asks why the conclusions of Trojan and Vlasov on the acoustics of a tachyon Fermi gas are incorrect.
In the second, the document is a 515-line report on start-ups, and the question asks how many new unicorns India, South Korea, and the UK added in Q3 CY22 and what share of all new unicorns this represents.
Both answers are judged correct.

\begin{tcolorbox}[casebox, fontupper=\scriptsize, fontlower=\scriptsize, title={\textbf{Two tasks under the evolved Faithfulness Verification harness}}]
\textbf{36-line document} \cmark\par
\texttt{scan()} $\to$ \textit{Total lines: 36}\par
\textit{``[\dots] This is a \good{SHORT document ($\le$120 lines)} [\dots] Since this is a short document, \good{I should read the full content with read\_lines(0, -1) first}''}\par
\good{\texttt{read\_lines(0, -1)}} $\to$ the whole document\par
\texttt{complete\_task}: \textit{``[\dots] They used an incorrect equation for pressure (equation 4) [\dots] The sound velocity ranges between $1/\sqrt{2}$ and $1/\sqrt{3}$, always remaining below the speed of light [\dots]''}
\tcblower
\textbf{515-line document} \cmark\par
\texttt{scan()} $\to$ \textit{Total lines: 515}\par
\textit{``[\dots] The scan shows it's 515 lines long, which is \good{medium length}. [\dots] Let me start with grep searches for relevant terms.''}\par
\good{\texttt{grep("unicorn")}} $\to$ 18 matches\par
\good{\texttt{read\_lines(274, 282)}} $\to$ \textit{``[\dots] only 20 unicorns globally in Q3 CY22 [\dots]''} and \textit{``[\dots] India,South Korea and the UK each added two.''}\par
\texttt{complete\_task}: \textit{``[\dots] for a combined total of 6 unicorns. This represents 30\% of the 20 new unicorns globally in Q3 CY22.''}
\end{tcolorbox}

EvoIn also reads short documents in full more often than the base model.
Under the seed harness, which has no such rule, it reads the whole document before any \texttt{grep} call on 88.3\% of the short-document evolution-test examples in these categories, compared with 66.4\% for the base model; on the remaining examples, the rates are 45.1\% and 24.3\%.
A document counts as short when it falls in the shortest size class of the rule of its category, and the count covers the ten categories whose evolution-test split contains such documents (470 examples).

\end{document}